\documentclass[twocolumn]{article}
\usepackage[T1]{fontenc}
\usepackage{preprint}

\AtBeginDocument{%
  }
\usepackage{natbib}
\usepackage{xcolor}
\usepackage{color}
\usepackage{wrapfig}
\usepackage{geometry}
\usepackage{graphicx}
\usepackage{overpic}
\usepackage{amsmath,amssymb}
\usepackage[colorlinks,linkcolor=red,citecolor=blue]{hyperref}
\usepackage{url}
\usepackage{enumitem}

\usepackage{booktabs}   
\usepackage{tabularx}   
\usepackage{array}      
\usepackage{ragged2e}   
\usepackage{graphicx}
\usepackage{multirow} 
\newtheorem{theorem}{Theorem}[section]
\newtheorem{proposition}[theorem]{Proposition}

\title{Projective Normal Fields: A Convex Optimization Method for Constructing Smooth UDFs}

\author{Jiayi Kong\\S-Lab\\Nanyang Technological University\\Singapore\And
Chen Zong\\College of Mathematics\\Nanjing University of Aeronautics and Astronautics\\China
\And Fei Hou\\Institute of Software\\Chinese Academy of Sciences\\China
\And Junhui Hou\\Department of Computer Science\\City University of Hong Kong\\China
\And Wenping Wang\\Department of Computer Science and Engineering\\Texas A\&M University\\USA
\And Ying He\thanks{Corresponding author: Y. He (yhe@ntu.edu.sg) }\\S-Lab\\Nanyang Technological University\\Singapore
}

\renewcommand{\shorttitle}{Projective Normal Fields}
\date{}
\begin{document}
\maketitle
\newcommand{\fix}{\marginpar{FIX}}
\newcommand{\new}{\marginpar{NEW}}

\begin{abstract}
Constructing a smooth approximation of an unsigned distance field (UDF) from a raw point cloud is challenging because the input provides neither surface connectivity nor consistently oriented normals. Methods that directly learn a scalar UDF must also handle its non-differentiability on the zero level set and weak supervision away from the samples, which can lead to unstable optimization and spatial artifacts. We introduce \emph{Projective Normal Fields (PNFs)}, an orientation-free representation and convex optimization framework for estimating bidirectional normals from point positions alone. Each normal axis is encoded by a rank-one projector, which is invariant to normal reversal. We relax the non-convex set of hard projectors to its convex hull: the symmetric positive-semidefinite matrices with unit trace. Each soft tensor defines a local quadratic distance model and retains the relative weights of candidate normal axes. We estimate a coherent PNF by combining local tangent-plane fitting, soft-PCA anchoring, and overlap regularization on a fixed neighborhood graph. With positive anchoring weights, the objective is strongly convex and admits a unique global minimizer. Principal eigenvectors provide bidirectional normals, while the corresponding eigengaps provide spectral confidence indicators. We use these indicators to select and weight directional sources for heat diffusion, followed by Poisson integration to construct a regularized UDF approximation. By separating local geometry estimation from scalar-field construction, PNF avoids directly fitting the non-differentiable UDF. Experiments demonstrate reduced sensitivity to neighborhood size, competitive reconstruction under noise and outliers, and improved accuracy near non-manifold junctions. The project page is available at \url{https://anonymous17777367.github.io/PNF-page/}.

\end{abstract}
\section{Introduction}
\label{sec:intro}

Implicit distance fields provide a compact and flexible representation of 3D geometry for reconstruction, generation, and geometric processing. Signed distance fields (SDFs) are particularly effective when a consistent inside--outside partition is available~\citep{park2019deepsdf,wang2023neural,NeuralPull}. For open, non-orientable, or non-manifold surfaces, however, such a partition may be undefined or unnecessarily restrictive. Unsigned distance fields (UDFs) avoid this requirement and accommodate a broader class of surface geometries.


Constructing a reliable UDF from raw point samples remains challenging. The input provides neither surface connectivity nor consistently oriented normals, and noise, outliers, or nearby surface sheets can make local geometry difficult to infer. Moreover, an ideal UDF is non-differentiable at the surface, while regions away from the samples receive little direct supervision. 
Methods that learn the scalar field directly must therefore approximate a non-smooth target while controlling its behavior in weakly constrained regions~\citep{Xu2024DEUDF}. These difficulties can lead to unstable optimization, inaccurate gradients, and spatial artifacts.

We adopt a geometry-first perspective: estimate local surface geometry in an orientation-free form before constructing the global scalar field. A tangent plane depends on a normal axis, not on which of its two directions is designated positive. This observation motivates \emph{Projective Normal Fields (PNFs)}, which represent bidirectional normals using sign-invariant tensors. Relaxing hard normal-axis projectors to soft tensors allows neighboring estimates to be optimized jointly while retaining ambiguity in the local directional evidence.

\begin{figure*}[!h]
    \centering  
    \includegraphics[width=\linewidth]{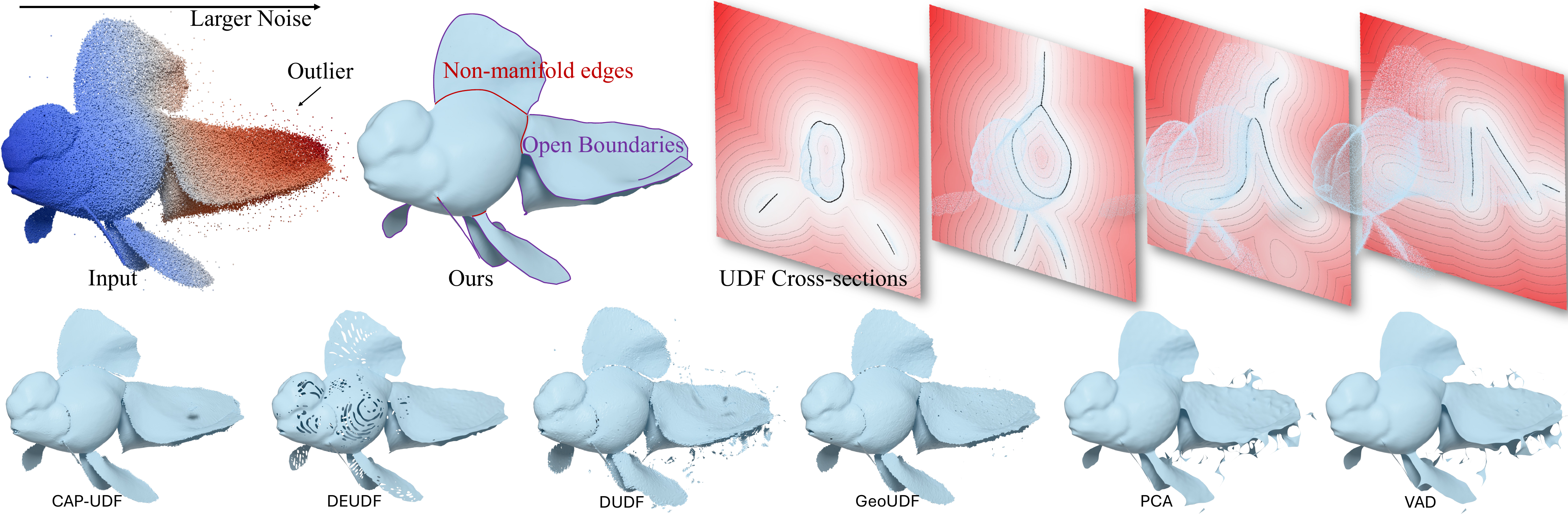}
    \caption{Reconstruction under spatially varying noise and outliers. Top row: the corrupted Goldfish point cloud, the PNF reconstruction, and planar cross-sections of the computed UDF. Warmer input point colors indicate greater
corruption. PNF preserves thin fins, open boundaries, and non-manifold junction geometry, while the color-coded distance values and level-set contours illustrate smooth spatial variation in the displayed slices. Bottom row: reconstructions from competing methods, which exhibit holes, surface irregularities, or spurious fragments.
}
    \label{fig:teaser}
\end{figure*}

Our method first estimates a coherent PNF from point positions through joint normal-axis optimization, then uses confidence-guided heat diffusion and scalar integration to construct a regularized UDF approximation. This two-stage design combines geometric evidence across samples and propagates directional information through the surrounding domain. Without requiring a globally consistent normal orientation, it accommodates open, non-orientable, and non-manifold surfaces. Our experiments evaluate sensitivity to neighborhood selection, robustness to noise and outliers, and reconstruction accuracy near non-manifold junctions.  Figure~\ref{fig:teaser} illustrates these capabilities on a corrupted Goldfish point cloud, alongside reconstructions from competing methods.

Our main contributions are threefold. First, we introduce Projective Normal Fields, an orientation-free representation based on hard normal-axis projectors and their convex relaxation to soft tensors. Second, we formulate PNF estimation as a strongly convex problem combining local tangent-plane fitting, soft-PCA anchoring, and graph-based overlap regularization, with a unique global minimizer under positive anchoring weights. Third, we develop a two-stage reconstruction framework that uses the decoded normal axes and their spectral confidence to select and weight directional sources for heat-based UDF construction.

\section{Related Work}
\label{sec:related-work}

\subsection{Direct UDF Learning} 
\label{subsec:directudflearning}
Direct approaches reconstruct surfaces by learning scalar unsigned-distance functions from point-cloud observations. NDF~\citep{chibane2020neural} introduces a neural UDF conditioned on sparse point clouds. GeoUDF~\citep{DBLP:conf/iccv/RenHCHW23} incorporates local geometry through learned combinations of distances to neighboring tangent planes. LoSF-UDF~\citep{losf-udf-2024} learns local UDFs from analytically defined shape functions, providing a lightweight model that generalizes across shapes. 

Projection-based objectives constrain distance-field learning through spatial query updates. Neural-Pull~\citep{NeuralPull}, formulated for SDFs, projects queries toward input points using predicted distances and gradients. In the unsigned setting, CAP-UDF~\citep{Zhou2022CAP-UDF} progressively pulls queries toward the surface under a field-consistency constraint, while   LevelSetUDF~\citep{Zhou2023LearningAM} improves zero-level-set continuity through projection from smoother nonzero level sets.
SuperUDF~\citep{10.1109/TVCG.2023.3318085} combines LOP-inspired projection with a learned geometry prior and sparsity regularization. 
RMSMS~\citep{liu2025unsupervised} uses recurrent multi-step query updates with distance and gradient regularization. These methods share the principle of enforcing consistency between the learned field, query motion, and sampled
surface geometry.

Another line of work addresses the non-differentiability of the ideal UDF at the surface by modifying the field representation or its regularization. DUDF~\citep{Fainstein2024DUDF} learns a differentiable hyperbolic transformation of the UDF. DEUDF~\citep{Xu2024DEUDF} combines an unconstrained neural output, normal alignment, adaptive Eikonal regularization, and a sinusoidal network to improve near-surface gradients and geometric detail. S$^2$DF \citep{yang2025monge} instead uses a differentiable scaled-squared distance representation with Monge--Amp\`ere regularization. These approaches retain a scalar UDF or a transformed distance field as the primary prediction target. 

Hybrid representations augment distance information with additional geometric or topological cues. HSDF~\citep{Wang22HSDF} jointly learns unsigned distances and an auxiliary sign field to support surface extraction. GIFS~\citep{ye2022gifs} predicts whether a surface separates a pair of spatial points and uses an auxiliary UDF branch to enhance spatial features. MPF~\citep{kong2026metricphase} separates an unsigned metric field from a smooth phase field, then combines them into a signed implicit function for thin-structure reconstruction.

\subsection{Geometry-First and Alternative Implicit Representations}
\label{subsec:indirectmethods}

Geometric reconstruction methods provide alternatives to direct distance-field regression. Moving least-squares surfaces~\citep{alexa2001point,kolluri2008provably} construct local surface approximations, while variational implicit point set methods~\citep{VIPSS,xia2025variational} recover implicit geometry through variational optimization. These methods provide geometric reconstruction precedents rather than direct neural UDF predictors. 

Closest-point and vector-valued representations offer another alternative by explicitly encoding the relationship between spatial queries and the surface.
The closest-surface-point representation~\citep{venkatesh2021csp} predicts the nearest surface point associated with each query, from which unsigned distance and local differential information can be derived. VF~\citep{mello2025neural} predicts the unit direction toward the nearest surface point, whereas  NVF~\citep{yang2023neural} predicts the full displacement vector, whose magnitude also encodes distance. These representations make
surface correspondence or directional information explicit, rather
than encoding geometry solely through scalar distance values.

Among geometry-first UDF construction methods, 
VAD~\citep{DBLP:journals/corr/abs-2510-12524} is the most closely related to PNF. It estimates bidirectional normals by reducing
discrepancies between local projection-distance fields and their
gradients across Voronoi bisectors. The optimized directional
information is then diffused through the ambient domain and integrated
into a scalar UDF. PNF retains this normal-first decomposition but
changes the normal-estimation formulation: it jointly optimizes soft
normal-axis tensors over a fixed neighborhood graph, without requiring
a Voronoi construction. With positive anchoring weights, the resulting
objective is strongly convex and admits a unique global minimizer.
The distinction therefore lies in the representation and optimization
of the local normal field, rather than in the use of diffusion for
downstream UDF construction. Appendix~\ref{subsec:relation-to-vad} provides a detailed comparison.

\section{Overview}
\label{sec:overview}


Let $\mathcal{X}=\{\mathbf{p}_i\}_{i=1}^{N}\subset\mathbb{R}^3$ be a raw point cloud representing an unknown surface
\(\mathcal{S}\), which may be open, non-orientable, or
non-manifold. Given only the point positions, our method constructs a
smooth approximation of its unsigned distance field
\(u_{\mathcal{S}}\) in two stages. 

Stage~I estimates a local normal axis at each input point without imposing a global orientation. Since \(\mathbf{n}_i\) and \(-\mathbf{n}_i\) describe the same
local tangent plane, we represent their common axis by the rank-one projector \(\mathbf P_i=\mathbf n_i\mathbf n_i^\top\). This representation removes the arbitrary sign choice. We call the collection 
\(\{\mathbf{P}_i\}_{i=1}^N\) a \emph{hard projective normal field}. Because hard projectors form a non-convex set, we relax them to
soft normal-axis tensors \(\{\mathbf{M}_i\}_{i=1}^{N}\) in a convex feasible set. We estimate the resulting \emph{soft projective normal
field} by combining local tangent-plane evidence with compatibility between neighboring local distance models. With a fixed graph and positive anchoring weights, the objective is strongly convex and has a unique global minimizer. After optimization, a principal eigenvector provides a hard normal axis, while the corresponding eigengap indicates how strongly the tensor favors that axis. 

Stage~II uses the decoded axes and their confidence values to construct a global UDF approximation. Confidence-based source selection and weighting modulate the directional information propagated by heat diffusion, and Poisson integration assembles this information into a scalar field. The pipeline thus separates orientation-free local geometry estimation from global distance-field construction. 

\begin{figure*}[htp]
    \centering
    \includegraphics[width=0.99\linewidth]{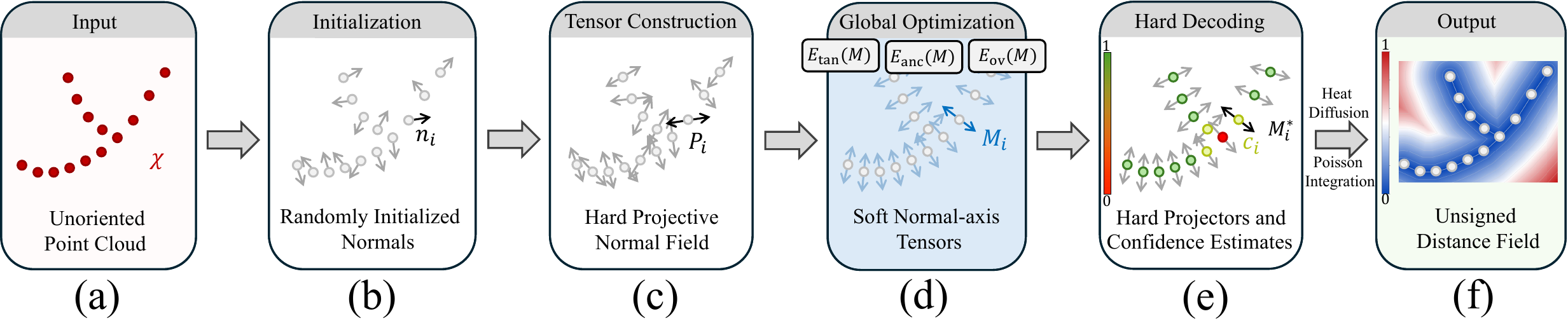}
    \vspace{-0.1in}
    \caption{Two-stage PNF reconstruction on a 2D Y-shaped point cloud. Given unoriented samples (a), Stage~I initializes random normal axes (b), encodes them as rank-one projectors (c), and optimizes their soft
relaxation using tangent-plane fitting, soft-PCA anchoring, and overlap regularization (d). Principal eigenvectors yield the decoded normal axes, while eigengaps provide confidence values (e). Bidirectional arrows represent unoriented axes; colors in (e) indicate confidence, which is
lower near the junction, where multiple branch directions compete, and higher along the regular branches. Stage~II applies confidence-guided heat diffusion followed by Poisson integration
to construct the UDF approximation (f).
    }
    \label{fig:pipeline}
\end{figure*}

Figure~\ref{fig:pipeline} illustrates the process on a 2D toy model. We present Stage~I in Section~\ref{sec:convex-optimization} and describe Stage~II in Appendix~\ref{sec:hm}.

\section{Convex Optimization of Bidirectional Normals}
\label{sec:convex-optimization}

This section develops the normal-axis representation and optimization used in Stage~I. We first describe hard normal-axis projectors, then introduce their convex relaxation to soft tensors, and finally formulate a graph-based objective for estimating a coherent projective normal field. 

\subsection{Hard Normal-Axis Projectors}
\label{subsec:hard-normal-axis-projectors}

Let $\mathbf n\in\mathbb S^2$ be a unit normal. The vectors $\mathbf n$ and $-\mathbf n$ represent the same unoriented normal axis $\operatorname{span}\{\mathbf n\}$, which we encode by
$\mathbf P=\mathbf n\mathbf n^\top$. We call $\mathbf P$ a \emph{hard normal-axis projector} or simply a \emph{hard projector}. This representation is sign invariant because
$(-\mathbf n)(-\mathbf n)^\top=\mathbf n\mathbf n^\top$. Conversely, every rank-one orthogonal projector in $\mathbb R^3$ has this form, with $\mathbf n$ unique up to sign. Hard projectors and bidirectional normals are therefore equivalent representations.

Since $\|\mathbf n\|_2=1$, the projector $\mathbf P$ satisfies $\mathbf P=\mathbf P^\top$, $\mathbf P\succeq\mathbf 0$, $\mathbf P^2=\mathbf P$, and $\operatorname{rank}(\mathbf P)=\operatorname{tr}(\mathbf P)=1$. For any displacement $\mathbf d\in\mathbb R^3$, $\mathbf P\mathbf d=(\mathbf n^\top\mathbf d)\mathbf n$ is its normal component, while $(\mathbf I-\mathbf P)\mathbf d$ is its tangential component. In particular,
$\|\mathbf P\mathbf d\|_2^2=\mathbf d^\top\mathbf P\mathbf d=(\mathbf n^\top\mathbf d)^2$. The tangent plane through $\mathbf p$ with normal projector $\mathbf{P}$ is 
$\Pi(\mathbf p,\mathbf P)
    :=\{\mathbf x\in\mathbb R^3:\mathbf P(\mathbf x-\mathbf p)=\mathbf 0\}
    =\{\mathbf x\in\mathbb R^3:(\mathbf x-\mathbf p)^\top
      \mathbf P(\mathbf x-\mathbf p)=0\}$.
For a query point $\mathbf y\in\mathbb R^3$, its squared Euclidean distance to this plane is $\operatorname{dist}^2\!\left(\mathbf y,\Pi(\mathbf p,\mathbf P)\right)
    =\|\mathbf P(\mathbf y-\mathbf p)\|_2^2
    =(\mathbf y-\mathbf p)^\top\mathbf P(\mathbf y-\mathbf p)$.

Consider two hard projectors $\mathbf P=\mathbf n\mathbf n^\top$ and
$\mathbf Q=\mathbf m\mathbf m^\top$, and let
$\theta:=\arccos|\mathbf n^\top\mathbf m|\in[0,\pi/2]$ be the acute angle between their axes. Then $\operatorname{tr}(\mathbf P\mathbf Q) =(\mathbf n^\top\mathbf m)^2=\cos^2\theta$, and $\frac12\|\mathbf P-\mathbf Q\|_F^2
    =1-(\mathbf n^\top\mathbf m)^2=\sin^2\theta$. The Frobenius distance therefore compares normal axes directly, without requiring sign alignment.


The set of all \emph{hard normal-axis projectors} is
\begin{equation}
    \mathcal P_3
    :=\{\mathbf n\mathbf n^\top:\mathbf n\in\mathbb S^2\}
    =\left\{
      \begin{aligned}
        &\mathbf P\in\operatorname{Sym}(3):\;
         \mathbf P\succeq\mathbf 0,\\
        &\operatorname{rank}(\mathbf P)=1,\;
         \operatorname{tr}(\mathbf P)=1
      \end{aligned}
    \right\},
    \label{eq:hard-projector-set}
\end{equation}
where $\operatorname{Sym}(3)$ denotes the space of real symmetric $3\times 3$ matrices. The term \emph{projective} reflects the identification of opposite unit normals 
\(\mathbf n\) and \(-\mathbf n\) as a single unoriented axis. Accordingly, $\mathcal{P}_3$ is a matrix realization of the real projective plane \(\mathbb{RP}^2=\mathbb{S}^2/\{\pm 1\}\).  A discrete hard PNF assigns one such projector to each input point. 

The projector representation also behaves naturally under sign changes: while the signed vectors $\mathbf n$ and $-\mathbf n$ cancel when averaged, their projectors are identical. An average of different hard projectors, however, is generally no longer rank one. This observation motivates the soft representation introduced next.

\subsection{Soft Normal-Axis Tensors}
\label{subsec:soft-normal-axis-tensors}

The hard-projector set $\mathcal P_3$ is non-convex. For example, the midpoint of the projectors associated with the orthogonal axes $(1,0,0)^\top$ and $(0,1,0)^\top$ is $\operatorname{diag}(1/2,1/2,0)$, which has rank two. We obtain a convex relaxation by dropping the rank-one constraint while retaining symmetry, positive semidefiniteness, and unit trace. Within this class, the rank-one and idempotence conditions are equivalent.

A \emph{soft normal-axis tensor} is a matrix in 
\begin{equation}
    \mathcal P_3^{\mathrm{cvx}}
    :=\{\mathbf M\in\operatorname{Sym}(3):
       \mathbf M\succeq\mathbf 0,
       \operatorname{tr}(\mathbf M)=1\}.
    \label{eq:soft-projector-set}
\end{equation}
For brevity, we also use the term \emph{soft projector}. This is only shorthand: a general soft tensor need not satisfy
$\mathbf M^2=\mathbf M$ and is therefore not an orthogonal projector.

\begin{proposition}[Convex relaxation and the hard--soft relationship]
\label{prop:soft-set-convex}
The set $\mathcal P_3^{\mathrm{cvx}}$ is compact and convex. Every
$\mathbf M\in\mathcal P_3^{\mathrm{cvx}}$ admits an eigendecomposition
\begin{equation}
    \mathbf M
    =\sum_{r=1}^{3}\lambda_r\mathbf e_r\mathbf e_r^\top,
    \qquad
    \lambda_r\ge0,
    \qquad
    \sum_{r=1}^{3}\lambda_r=1,
    \label{eq:soft-spectral-mixture}
\end{equation}
where $\{\mathbf e_r\}_{r=1}^3$ is an orthonormal basis and each
$\mathbf e_r\mathbf e_r^\top$ is a hard projector. Conversely, every convex combination of hard projectors belongs to
$\mathcal P_3^{\mathrm{cvx}}$. Hence, $\mathcal P_3^{\mathrm{cvx}}$ is the convex hull of $\mathcal P_3$, and its extreme points are exactly the hard projectors.
\end{proposition}

Appendix~\ref{app:soft-set-proof} provides the proof. The eigendecomposition gives a direct geometric interpretation: the eigenvectors define candidate normal axes, and the eigenvalues specify their nonnegative mixture weights. A hard projector assigns all weight to one axis, whereas a more diffuse spectrum retains competing directional preferences.  We optimize the soft tensors first and decode a single hard normal axis afterward.  

\begin{proposition}[Nearest hard projector]
\label{prop:nearest-hard-projector}
Let $\lambda_1\ge\lambda_2\ge\lambda_3$ be the eigenvalues of
$\mathbf M\in\mathcal P_3^{\mathrm{cvx}}$, and let $\mathbf{e}_1$ be any unit eigenvector associated with $\lambda_1$. Define the hardening operator $\operatorname{Hard}(\mathbf M):=\mathbf e_1\mathbf e_1^\top$. Then
\begin{equation}
    \operatorname{Hard}(\mathbf M)
    \in\arg\min_{\mathbf P\in\mathcal P_3}
       \|\mathbf M-\mathbf P\|_F^2,
    \label{eq:nearest-hard-projector}
\end{equation} with minimum value
$1+\operatorname{tr}(\mathbf M^2)-2\lambda_1$.
The minimizing projector is unique if and only if $\lambda_1>\lambda_2$.
Moreover, $\frac13\le\operatorname{tr}(\mathbf M^2)\le1$, with equality at the upper bound exactly for hard projectors and at the lower bound exactly for the isotropic tensor $\mathbf I/3$.
\end{proposition}

Appendix~\ref{app:nearest-hard-proof} provides the proof. The soft representation offers two advantages: it removes the non-convex rank constraint from optimization, and it retains directional preference in the tensor spectrum. We use the principal 
eigengap
$c(\mathbf M):=\lambda_1-\lambda_2$ as a spectral confidence indicator. A large gap indicates a clearly preferred axis, whereas a small gap indicates competing leading directions. This indicator describes the optimized tensor's preference, not the correctness of the decoded normal.

\subsection{A Convex PNF Objective}
\label{subsec:convex-pnf-objective}

\noindent\textbf{Neighborhood graph.}
We optimize one soft tensor
\(\mathbf M_i\in\mathcal P_3^{\mathrm{cvx}}\)
at each input point \(\mathbf p_i\). To couple neighboring estimates, we
construct a fixed undirected graph \(G=(V,E)\), where
\(V=\{1,\ldots,N\}\). Possible constructions include radius-based neighborhoods, symmetrized \(k\)-nearest neighbors, and Voronoi adjacency. The graph is 
computed before optimization and remains fixed during the solve. Each edge expresses the intended compatibility between local surface models at its endpoints. For every edge, we define the unit direction $\widehat{\mathbf d}_{ij}
:=
(\mathbf p_j-\mathbf p_i)/\|\mathbf p_j-\mathbf p_i\|_2$. Either orientation may be used for an undirected edge: replacing
\(\widehat{\mathbf d}_{ij}\) by
\(-\widehat{\mathbf d}_{ij}\) leaves the objective
unchanged.

\noindent\textbf{Local scatter matrix.}
For each point \(\mathbf p_i\), let \(\mathcal N_C(i)\) be a
fixed, nonempty neighborhood. We define $\mathbf C_i
:=
\frac{1}{|\mathcal N_C(i)|}
\sum_{j\in\mathcal N_C(i)}
(\mathbf p_j-\mathbf p_i)
(\mathbf p_j-\mathbf p_i)^\top$.
The symmetric positive-semidefinite matrix \(\mathbf C_i\) describes the directional spread of the neighboring points around
\(\mathbf p_i\). For any unit vector \(\mathbf v\),
$\mathbf v^\top\mathbf C_i\mathbf v
=
\frac{1}{|\mathcal N_C(i)|}
\sum_{j\in\mathcal N_C(i)}
\bigl(
\mathbf v^\top(\mathbf p_j-\mathbf p_i)
\bigr)^2$, which is the average squared
displacement along \(\mathbf v\).
For a locally planar neighborhood, the direction of least spread approximates the surface normal.  


\noindent\textbf{Soft-PCA prior.}
We convert the local scatter matrix into a fixed soft prior
$\overline{\mathbf M}_i
:=
\frac{
\exp(-\mathbf C_i/\tau_i)
}{
\operatorname{tr}
\bigl(
\exp(-\mathbf C_i/\tau_i)
\bigr)
}
\in\mathcal P_3^{\mathrm{cvx}}$, with $\tau_i>0$. The matrix exponential preserves the eigenvectors of \(\mathbf C_i\)
and assigns larger weights to directions with smaller eigenvalues. The prior therefore favors the local PCA normal while retaining competing directions when the local geometry is ambiguous. The parameter \(\tau_i\) controls this preference: as
\(\tau_i\to 0\), the prior approaches the rank-one PCA normal projector provided that the smallest eigenvalue of \(\mathbf C_i\) is simple.

\noindent\textbf{Local quadratic model.}
Each point--tensor pair \((\mathbf p_i,\mathbf M_i)\) defines a local model of squared distance, 
$\psi_i(\mathbf x;\mathbf M_i)
:=
(\mathbf x-\mathbf p_i)^\top
\mathbf M_i
(\mathbf x-\mathbf p_i)$. For a hard projector \(\mathbf P_i=\mathbf n_i\mathbf n_i^\top\), this becomes $\psi_i(\mathbf x;\mathbf P_i)
=
\bigl(
\mathbf n_i^\top(\mathbf x-\mathbf p_i)
\bigr)^2$, the squared distance to the plane
through \(\mathbf p_i\) with normal axis \(\mathbf n_i\). For a soft
tensor, it is a weighted blend of squared
distances to planes associated with the tensor's candidate axes. 
Although \(\psi_i\) is quadratic in \(\mathbf x\), it is linear in the unknown tensor \(\mathbf M_i\).

\noindent\textbf{Tangent-plane fitting.} We collect the unknowns as $\mathbf M:=(\mathbf M_1,\ldots,\mathbf M_N)$ and fit each local quadratic model to the neighboring
input points. By the definition of $\mathbf{C}_i$, 
$\operatorname{tr}(\mathbf C_i\mathbf M_i)
=
\frac{1}{|\mathcal N_C(i)|
}
\sum_{j\in\mathcal N_C(i)}
(\mathbf p_j-\mathbf p_i)^\top
\mathbf M_i
(\mathbf p_j-\mathbf p_i)
=\frac{1}{|\mathcal N_C(i)|}
\sum_{j\in\mathcal N_C(i)}
\psi_i(\mathbf p_j;\mathbf M_i)$.
The trace is therefore the 
average value assigned by the local model to its neighboring
samples. This motivates
$E_{\mathrm{tan}}(\mathbf M)
:=
\sum_{i=1}^{N}
\gamma_i
\operatorname{tr}(\mathbf C_i\mathbf M_i)$, where \(\gamma_i\geq0\) is fixed.
For a hard projector
\(\mathbf P_i\), the trace
becomes
$\operatorname{tr}(\mathbf C_i\mathbf P_i)
=
\frac{1}{
|\mathcal N_C(i)|}
\sum_{j\in\mathcal N_C(i)}
\bigl(
\mathbf n_i^\top(\mathbf p_j-\mathbf p_i)
\bigr)^2$. Each summand is the squared distance of a neighboring point to the estimated plane. Minimizing \(E_{\mathrm{tan}}\) therefore encourages local tangent-plane agreement. The energy is linear in the unknown tensors.

\noindent\textbf{Anchoring.} Tangent-plane fitting favors directions of small local
variation, but does not by itself retain the directional balance encoded by the soft-PCA prior. We therefore penalize deviations from that prior $E_{\mathrm{anc}}(\mathbf M)
:=
\frac{1}{2}
\sum_{i=1}^{N}
\rho_i
\left\|
\mathbf M_i-\overline{\mathbf M}_i
\right\|_F^2$, where \(\rho_i\geq0\) is a fixed anchoring weight. A larger
\(\rho_i\) places greater emphasis on the local soft-PCA estimate, whereas
a smaller value allows stronger modification through neighboring evidence. The anchor term thus balances local fidelity against graph-based coupling. Strictly positive anchoring weights also make the quadratic anchoring energy strongly convex. 

\noindent\textbf{Overlap regularization.}
The fitting and anchoring terms act independently at each
sample. To obtain a coherent field, we additionally encourage neighboring 
quadratic models to agree where their local neighborhoods overlap. Differentiating the local model gives $
\nabla_{\mathbf x}\psi_i(\mathbf x;\mathbf M_i)
=
2\mathbf M_i(\mathbf x-\mathbf p_i)$, and 
 $\nabla_{\mathbf x}^2\psi_i(\mathbf x;\mathbf M_i)
=
2\mathbf M_i$. Thus, 
\(\mathbf M_i-\mathbf M_j\) is proportional to the Hessian discrepancy between
two neighboring models. At the midpoint $\mathbf{q}_{ij}=(\mathbf p_i+\mathbf p_j)/2$ of an edge \(\{i,j\}\in E\), the gradient difference is 
$\nabla_{\mathbf x}
\psi_i
\left(
\mathbf{q}_{ij};
\mathbf M_i
\right)
-
\nabla_{\mathbf x}
\psi_j
\left(
\mathbf{q}_{ij};
\mathbf M_j
\right)
=
\|\mathbf p_j-\mathbf p_i\|_2
(\mathbf M_i+\mathbf M_j)
\widehat{\mathbf d}_{ij}$,
whereas their value difference is
$\psi_i
\left(
\mathbf{q}_{ij};
\mathbf M_i
\right)
-
\psi_j
\left(
\mathbf{q}_{ij};
\mathbf M_j
\right)
=
0.25\|\mathbf p_j-\mathbf p_i\|_2^2
\widehat{\mathbf d}_{ij}^{\top}
(\mathbf M_i-\mathbf M_j)
\widehat{\mathbf d}_{ij}$. These identities motivate the overlap energy
$E_{\mathrm{ov}}(\mathbf M)
:=
\frac{1}{2}
\sum_{\{i,j\}\in E}
w_{ij}
\Big[
\beta_s
\|\mathbf M_i-\mathbf M_j\|_F^2
+
\beta_t
\|(\mathbf M_i+\mathbf M_j)
\widehat{\mathbf d}_{ij}\|_2^2
+
\beta_q
\Big(
\widehat{\mathbf d}_{ij}^{\top}
(\mathbf M_i-\mathbf M_j)
\widehat{\mathbf d}_{ij}
\Big)^2
\Big]$, where \(w_{ij}\geq 0\) is a fixed symmetric edge weight, and \(\beta_s,\beta_t,\beta_q\geq0\) are fixed global coefficients. We consider uniform weights $w_{ij}\equiv 1$ or Gaussian distance-decaying weights $w_{ij}=\operatorname{exp}\left(-\|\mathbf{p}_i-\mathbf{p}_j\|^2/(2\sigma^2)\right)$, where $\sigma>0$ is a fixed global length scale. The three terms penalize discrepancies in Hessians, midpoint gradients, and midpoint values, respectively. Gradient differences are normalized by the edge length, and value differences by its square; fixed proportionality factors are absorbed into the global coefficients. This normalization prevents the residuals from increasing solely because an edge is longer, while the edge weights modulate the influence of individual neighbors. 


\paragraph{PNF objective.}
Combining the three terms, we define
$E_{\mathrm{PNF}}(\mathbf M)
:=
E_{\mathrm{tan}}(\mathbf M)
+
E_{\mathrm{anc}}(\mathbf M)
+
E_{\mathrm{ov}}(\mathbf M)$,
and solve
\begin{equation}
\mathbf M^\star
\in
\arg\min_{\mathbf M\in(\mathcal P_3^{\mathrm{cvx}})^N}
E_{\mathrm{PNF}}(\mathbf M).
\label{eqn:m-star}
\end{equation}
All neighborhoods, priors, edge directions, and weights are fixed
before optimization. The fitting term uses local point positions,
the anchor term retains soft-PCA evidence, and the overlap term
couples neighboring local models.


\begin{theorem}[Convexity and uniqueness of PNF optimization]
\label{thm:pnf-convexity} Under the fixed-data construction above, assume that $\rho_i>0$ for every $i$, and let $\rho_{\min}:=\min_{1\leq i\leq N}\rho_i>0$. Then \(E_{\mathrm{PNF}}\) is continuous and $\rho_{\min}$-strongly convex on \((\mathcal P_3^{\mathrm{cvx}})^N\) with respect to the block Frobenius norm
$\|\mathbf M\|_{\mathbb F}^2 := \sum_{i=1}^{N}\|\mathbf M_i\|_F^2$. Consequently, the constrained minimization problem in Equation~(\ref{eqn:m-star})
has a unique global minimizer.
\end{theorem}

Appendix~\ref{app:objective-convexity-proof} provides the proof. This guarantee concerns the selected fixed-graph model; it does not
ensure that every retained edge connects geometrically compatible
surface samples.

\noindent\textbf{Hard bidirectional normal decoding.} From the optimized tensors, we recover hard normal projectors $\mathbf P_i
:=
\operatorname{Hard}(\mathbf M_i^\star)$ and spectral confidence values $c_i
:=
\lambda_{i1}-\lambda_{i2}$,
where
\(\lambda_{i1}\geq\lambda_{i2}\geq\lambda_{i3}\)
are the eigenvalues of \(\mathbf M_i^\star\). The eigengap indicates
how distinctly the optimized tensor selects its leading axis. Stage~II uses this confidence to modulate the axis's contribution to directional diffusion.

\begin{table*}[h]
\centering
\caption{Sensitivity to neighborhood size \(k\).
We report mean absolute cosine similarity between estimated  and reference normal axes on four models with thin structures. Higher values are better, with \(1\) indicating
perfect agreement. PNF exhibits less variation across the tested neighborhood sizes than local PCA. The better result for each model and neighborhood size is shown in
\textbf{bold}. The plots visualize the same data.}
\label{tab:k-sensitivity}
\begin{minipage}[c]{0.58\textwidth}
\centering
\small
\setlength{\tabcolsep}{2pt}
\renewcommand{\arraystretch}{0.950}
\begin{small}
\begin{tabular}{ll cccccc}
\toprule
\textbf{Model} & \textbf{Method} & $k=3$ & $k=5$ & $k=10$ & $k=15$ & $k=20$ & $k=30$ \\
\midrule
\multirow{2}{*}{Toy} & PCA  & 0.9693 & 0.9900 & 0.9660 & 0.9746 & 0.9828 & 0.9833 \\
                       & Ours & \textbf{0.9934} & \textbf{0.9937} & \textbf{0.9938} & \textbf{0.9937} & \textbf{0.9937} & \textbf{0.9934} \\
\midrule
\multirow{2}{*}{Ship}  & PCA  & 0.9525 & 0.9079 & 0.8815 & 0.9301 & 0.9408 & 0.9451 \\
                       & Ours & \textbf{0.9628} & \textbf{0.9621} & \textbf{0.9555} & \textbf{0.9523} & \textbf{0.9505} & \textbf{0.9515} \\
\midrule
\multirow{2}{*}{Leaf}  & PCA  & 0.9609 & 0.9725 & 0.9707 & 0.9605 & 0.9629 & 0.9778 \\
                       & Ours & \textbf{0.9863} & \textbf{0.9878} & \textbf{0.9876} & \textbf{0.9869} & \textbf{0.9860} & \textbf{0.9847} \\
\midrule
\multirow{2}{*}{Coil}  & PCA  & 0.9000 & 0.9576 & 0.9733 & 0.9755 & \textbf{0.9763} & 0.9750 \\
                       & Ours & \textbf{0.9700} & \textbf{0.9795} & \textbf{0.9796} & \textbf{0.9779} & 0.9757 & \textbf{0.9841} \\
\bottomrule
\end{tabular}
\end{small}
\end{minipage}
\hfill
\begin{minipage}[c]{0.4\textwidth}
\centering
\includegraphics[width=\linewidth]{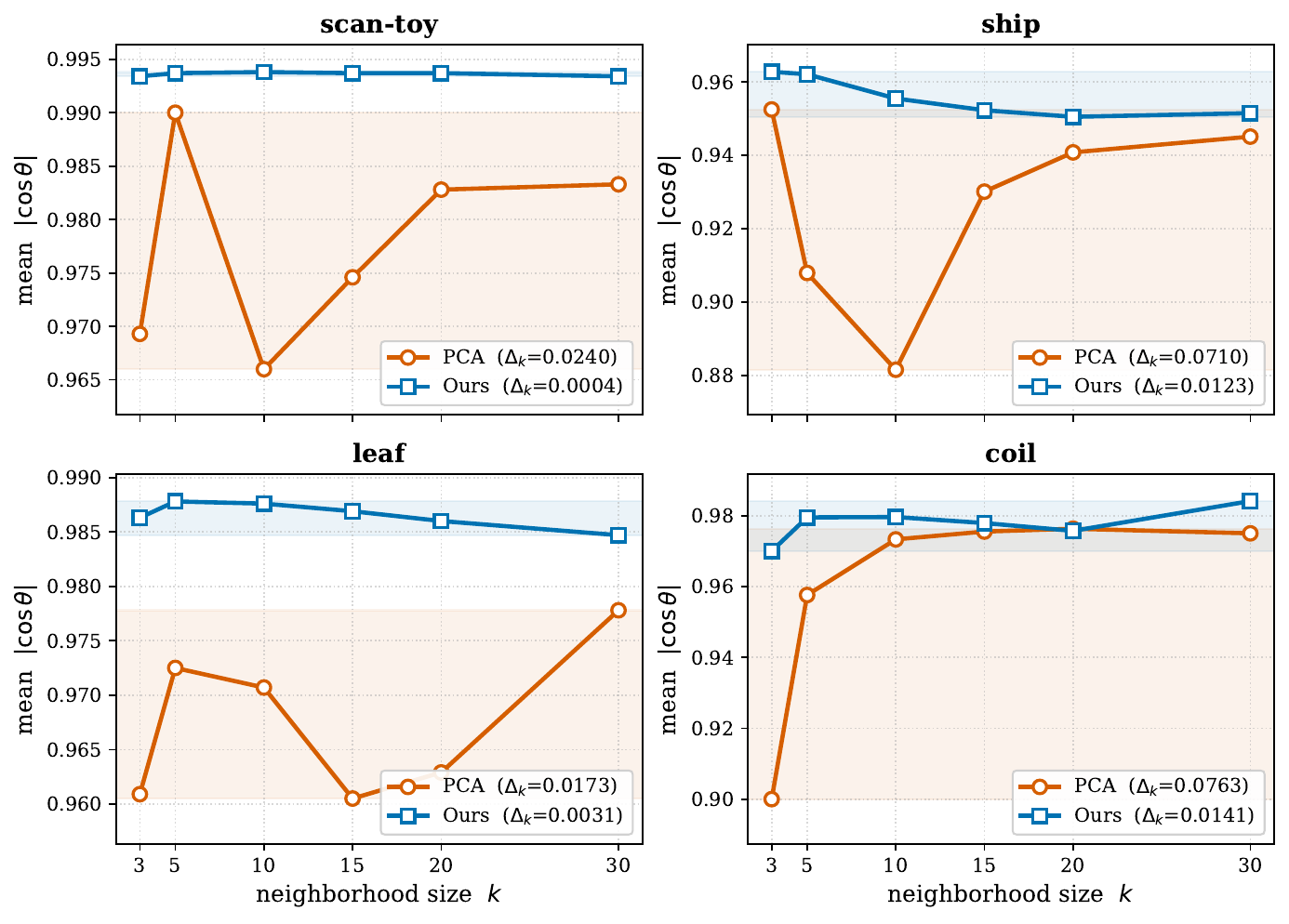}
\end{minipage}
\end{table*}

\section{Experimental Results}
\label{sec:results}

\begin{figure*}[!htbp]
    \centering
    \IfFileExists{pic/pnf-pca.pdf}{%
        \includegraphics[width=\linewidth]{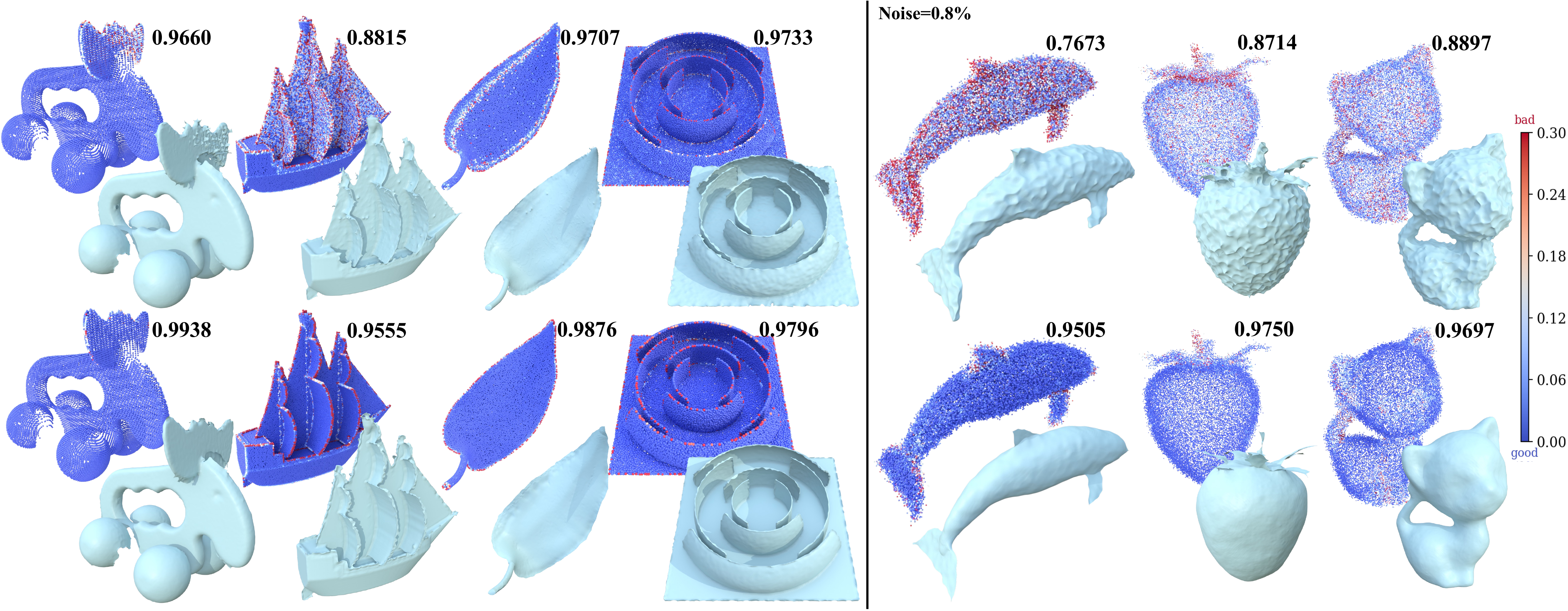}\\
        \makebox[0.9in]{Toy}
        \makebox[0.86in]{Ship}
        \makebox[0.9in]{Leaf}
        \makebox[0.9in]{Coil}
        \makebox[0.4in]{}
        \makebox[0.9in]{Dolphin}
        \makebox[0.9in]{Strawberry}
        \makebox[0.9in]{Kitten}%
    }{%
        \fbox{\parbox[c][3cm][c]{0.94\linewidth}{\centering Normal-axis comparison\\[6pt]\small Visualization pending}}%
    }
    \vspace{-0.1in}
    \caption{Global PNF optimization vs. local PCA fitting. We compare normal-axis estimates from PCA (top row) and PNF (bottom row) on thin structures with $k=10$ (left) and inputs corrupted by $0.8\%$ noise (right), together with the corresponding surface reconstruction results. Colors indicate per-point normal error, from blue (low) to red (high). Numbers report the mean absolute cosine similarity between estimated and reference normals; higher values are better, with $1$ indicating perfect agreement. Unlike PCA, which fits each neighborhood independently, PNF jointly optimizes the normal-axis field over the entire neighborhood graph. This global formulation combines local geometric evidence with inter-sample consistency, yielding more accurate axes and improved surface reconstructions in these challenging examples.}
    \label{fig:knn-noise-vis}
\end{figure*}
\noindent\textbf{PCA versus PNF: ambiguous local neighborhoods.} PCA estimates a normal axis from the direction of least local point variation. This estimate becomes less reliable when the neighborhood lacks a clear tangent-plane structure. Such ambiguity often arises in three
challenging settings: noisy inputs, where perturbations obscure local directional structure; outlier-contaminated inputs, where spurious samples distort the local point distribution; and thin structures, where a neighborhood may contain points from distinct nearby surface sheets. In each case, the neighborhood size \(k\) can substantially affect the PCA estimate. Rather than estimating each axis independently, PNF jointly optimizes the normal-axis field using tangent-plane evidence, soft-PCA anchoring, and compatibility between neighboring quadratic models. 
Figure~\ref{fig:knn-noise-vis} shows higher normal-axis accuracy for PNF on the illustrated thin structures and noisy inputs.
Table~\ref{tab:k-sensitivity} further shows smaller variations in accuracy across the tested neighborhood sizes. These results support the benefit of joint normal-axis optimization when individual neighborhoods provide ambiguous geometric evidence.

\noindent\textbf{Confidence as a geometric cue.}
\label{par:confidence-applications}
Besides a decoded normal axis, PNF provides the eigengap confidence $c_i$. This quantity measures how strongly the optimized tensor favors its leading axis. During reconstruction, we use confidence to modulate directional source contributions and suppress sources below a fixed threshold (Appendix~\ref{sec:hm}). This reduces the influence of weakly preferred axes without automatically discarding the corresponding input samples. Confidence also provides a cue for locating ambiguous normal axes before surface reconstruction. Figure~\ref{fig:confidence-applications} visualizes the optimized confidence on non-manifold geometry and outlier-contaminated inputs. Low values can occur near junctions with competing axes or around samples with inconsistent directional support. However, confidence is not a point-type classifier: low values alone cannot distinguish junctions, outliers, noise, or sparse sampling, and high values do not guarantee a correct normal.


\noindent\textbf{Comparisons.}
We compare PNF with CAP-UDF~\citep{Zhou2022CAP-UDF}, GeoUDF~\citep{DBLP:conf/iccv/RenHCHW23}, DUDF~\citep{Fainstein2024DUDF}, DEUDF~\citep{Xu2024DEUDF}, and VAD~\citep{DBLP:journals/corr/abs-2510-12524}. For the non-manifold experiments, we also include PCA+HM, which replaces PNF estimation with local PCA normal axes~\citep{hoppe1992surface} before heat-based UDF construction. The experiments examine robustness to corrupted inputs and reconstruction near non-manifold junctions, where a unique normal axis may be undefined. 

\noindent\textbf{Benchmark and evaluation.} The benchmark contains 60 models evaluated under five input conditions: clean samples, two noise levels ($0.3\%$ and $0.8\%$), and two outlier levels ($2\%$ and $5\%$). The underlying model set is the same across conditions. We use ground-truth meshes as references when available and otherwise use the original clean point clouds. Table~\ref{tab:overall-benchmark} reports arithmetic means of the per-model errors. We evaluate directed Chamfer distance (CD) and Hausdorff distance (HD) from the reference geometry to the reconstruction, corresponding to the mean and maximum nearest-surface distances, respectively. PNF achieves lower CD and HD than VAD under all five tested conditions. Across all compared methods, PNF obtains the lowest CD in four conditions and the second-lowest CD under $0.8\%$ noise. GeoUDF obtains the lowest HD in four conditions, while PNF obtains the lowest HD under \(0.8\%\) noise. These results demonstrate competitive performance of the complete PNF pipeline under the evaluated input corruptions.

\begin{figure*}[t]
    \centering
    \includegraphics[width=\linewidth]{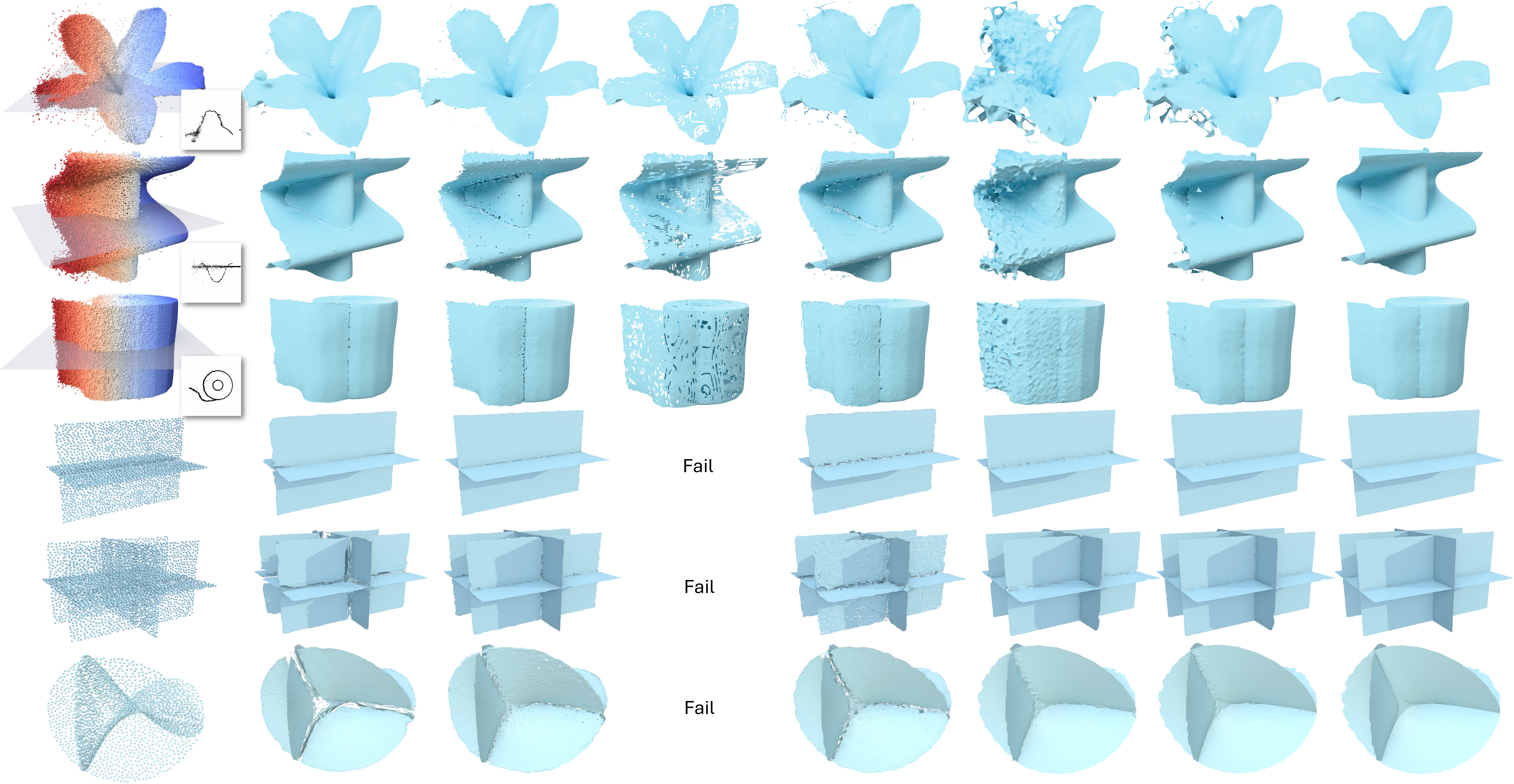}
    \makebox[0.13\linewidth][c]{Input}
         \makebox[0.11\linewidth][c]{CAP-UDF} 
     \makebox[0.115\linewidth][c]{GeoUDF}
       \makebox[0.115\linewidth][c]{DEUDF}       \makebox[0.115\linewidth][c]{DUDF}               \makebox[0.115\linewidth][c]{PCA+HM} 
          \makebox[0.11\linewidth][c]{VAD} 
     \makebox[0.1\linewidth][c]{Ours}
     \vspace{-0.1in}
\caption{Qualitative comparison on non-manifold models. The top three rows show reconstructions from inputs with spatially varying noise and outliers; warmer point colors indicate greater corruption, and insets show cross-sections at the indicated planes. The bottom three rows show a cross-junction, a multi-junction, and the self-intersecting Henneberg surface. Several baselines exhibit holes, surface irregularities, or spurious floating fragments, whereas PNF  produces smoother, more coherent reconstructions that closely follow the junction geometry. DEUDF failed on the three synthetic non-manifold models due to its reliance on locally estimated PCA normals for gradient alignment. See also Table~\ref{tab:nonmanifold-eval} for quantitative results for these non-manifold models. }
    \label{fig:noise+outlier}
\end{figure*}

\noindent\textbf{Non-manifold reconstruction.} We further evaluate three synthetic models: a cross-junction, a multi-junction, and the self-intersecting Henneberg surface. The bottom three rows of Figure~\ref{fig:noise+outlier} show these models, which test reconstruction near locations where a single normal axis is insufficient to describe the local geometry. PNF retains competing directional preferences in its soft tensors and reduces the influence of low-confidence directional sources during Stage~II. Table~\ref{tab:nonmanifold-eval} reports global and junction-region accuracy. PNF achieves the lowest global CD, junction-region CD and HD95 on all three models, together with $100\%$ junction recall in each case, including ties with other methods. These results demonstrate accurate reconstruction near the evaluated non-manifold structures. Appendix~\ref{app:reconstruction-ablations} examines the effects of the normal estimator and confidence weighting and describes the graph-construction comparison.

\begin{table*}[t]
\centering
\caption{Robustness to input corruptions. Mean per-shape directed CD and HD, evaluated from the reference geometry to the reconstruction, on the same 60 test shapes. Distances are reported in units of \(10^{-3}\); lower values are better. The best and second-best results in each
metric column are shown in \textbf{bold} and \underline{underlined},
respectively.}
\label{tab:overall-benchmark}

\small
\setlength{\tabcolsep}{3pt}
\renewcommand{\arraystretch}{1.00}

\begin{tabular}{@{}l*{10}{r}@{}}
\toprule
\textbf{Method}
& \multicolumn{2}{c}{\textbf{Clean}}
& \multicolumn{2}{c}{\textbf{Noise (0.3\%)}}
& \multicolumn{2}{c}{\textbf{Noise (0.8\%)}}
& \multicolumn{2}{c}{\textbf{Outliers (2\%)}}
& \multicolumn{2}{c}{\textbf{Outliers (5\%)}} \\
\cmidrule(lr){2-3}
\cmidrule(lr){4-5}
\cmidrule(lr){6-7}
\cmidrule(lr){8-9}
\cmidrule(lr){10-11}
& \textbf{CD} & \textbf{HD}
& \textbf{CD} & \textbf{HD}
& \textbf{CD} & \textbf{HD}
& \textbf{CD} & \textbf{HD}
& \textbf{CD} & \textbf{HD} \\
\midrule

CAP-UDF
& 0.329 & \underline{6.592}
& 1.232 & 9.329
& 2.990 & 19.492
& 0.407 & 7.101
& 0.452 & \underline{7.886} \\

GeoUDF
& 0.206 & \textbf{5.253}
& 1.691 & \textbf{7.628}
& 2.968 & 11.888
& \underline{0.241} & \textbf{5.757}
& \underline{0.243} & \textbf{5.832} \\

DUDF
& 0.395 & 7.818
& 2.836 & 18.463
& \textbf{2.047} & 10.802
& 0.628 & 9.381
& 0.639 & 10.956 \\

DEUDF
& 0.465 & 15.079
& 2.358 & 15.999
& 4.634 & 24.952
& 0.615 & 19.658
& 0.671 & 21.009 \\

VAD
& \underline{0.133} & 7.195
& \underline{1.199} & 8.495
& 2.311 & \underline{10.491}
& 1.089 & 14.194
& 1.050 & 13.331 \\

\textbf{PNF (Ours)}
& \textbf{0.123} & 7.166
& \textbf{1.179} & \underline{8.377}
& \underline{2.199} & \textbf{10.172}
& \textbf{0.194} & \underline{6.707}
& \textbf{0.150} & 11.269 \\

\bottomrule
\end{tabular}
\end{table*}

\section{Conclusion}
\label{sec:conclusion}

We introduced Projective Normal Fields, an orientation-free representation
that separates local normal-axis estimation from global UDF construction.
By relaxing rank-one projectors to positive-semidefinite, unit-trace
tensors, we combine tangent-plane evidence, soft-PCA anchoring, and
overlap consistency in a strongly convex optimization on a fixed graph
with positive anchoring weights. The optimized tensors provide normal
axes and spectral confidence, which guide directional-source selection,
heat diffusion, and subsequent UDF construction. Experiments demonstrate
reduced sensitivity to neighborhood size compared with local PCA,
competitive reconstruction under noise and outliers, and improved
accuracy near the evaluated non-manifold junctions.

\bibliographystyle{ACM-Reference-Format}
\bibliography{bib}

@article{DBLP:journals/corr/abs-2510-12524,
  author       = {Jiayi Kong and
                  Chen Zong and
                  Junkai Deng and
                  Xuhui Chen and
                  Fei Hou and
                  Shiqing Xin and
                  Junhui Hou and
                  Chen Qian and
                  Ying He},
  title        = {Voronoi-Assisted Diffusion for Computing Unsigned Distance Fields
                  from Unoriented Points},
  journal      = {CoRR},
  volume       = {abs/2510.12524},
  year         = {2025},
}

@InProceedings{PUGeo,
author={Qian, Yue
and Hou, Junhui
and Kwong, Sam
and He, Ying},
title={PUGeo-Net: A Geometry-Centric Network for 3D Point Cloud Upsampling},
booktitle={ECCV 2020},
year={2020},
pages={752--769},
}

@article{feng2024heat,
  title={A heat method for generalized signed distance},
  author={Feng, Nicole and Crane, Keenan},
  journal={ACM Transactions on Graphics (TOG)},
  volume={43},
  number={4},
  pages={1--19},
  year={2024},
  publisher={ACM New York, NY, USA}
}

@inproceedings{Zhou2022CAP-UDF,
    title = {Learning Consistency-Aware Unsigned Distance Functions Progressively from Raw Point Clouds},
    author = {Zhou, Junsheng and Ma, Baorui and Liu, Yu-Shen and Fang, Yi and Han, Zhizhong},
    booktitle = {Advances in Neural Information Processing Systems (NeurIPS)},
    year = {2022}
}

@inproceedings{Xu2024DEUDF,
    author={Xu, Cheng and Hou, Fei and Wang, Wencheng and Qin, Hong and Zhang, Zhebin and He, Ying},
    title={Details Enhancement in Unsigned Distance Field Learning for High-fidelity 3D Surface Reconstruction},
    booktitle = {Proc. of AAAI},
    year = {2025},
}

@article{Zhou2023LearningAM,
  title={Learning a More Continuous Zero Level Set in Unsigned Distance Fields through Level Set Projection},
  author={Junsheng Zhou and Baorui Ma and Shujuan Li and Yu-Shen Liu and Zhizhong Han},
  journal={2023 IEEE/CVF International Conference on Computer Vision (ICCV)},
  year={2023},
  pages={3158-3169},
}

@inproceedings{hoppe1992surface,
    author = {Hoppe, Hugues and DeRose, Tony and Duchamp, Tom and McDonald, John and Stuetzle, Werner},
    title = {Surface reconstruction from unorganized points},
    year = {1992},
    isbn = {0897914791},
    publisher = {Association for Computing Machinery},
    address = {New York, NY, USA},
    booktitle = {Proceedings of the 19th Annual Conference on Computer Graphics and Interactive Techniques},
    pages = {71--78},
    numpages = {8},
    series = {SIGGRAPH '92}
}

@inproceedings{losf-udf-2024,
            author={Hu, Jiangbei and Li, Yanggeng and Hou, Fei and Hou, Junhui and Zhang, Zhebin and Wang, Shengfa and Lei, Na and He, Ying},
    title = {A Lightweight UDF Learning Framework for 3D Reconstruction Based on Local Shape Functions},
    booktitle = {{IEEE/CVF} Conference on Computer Vision and Pattern Recognition,
                  {CVPR} 2025},
    year = {2025}
}

@inproceedings{park2019deepsdf,
  author={Park, Jeong Joon and Florence, Peter and Straub, Julian and Newcombe, Richard and Lovegrove, Steven},
  booktitle={2019 IEEE/CVF Conference on Computer Vision and Pattern Recognition (CVPR)}, 
  title={DeepSDF: Learning Continuous Signed Distance Functions for Shape Representation}, 
  year={2019},
  volume={},
  number={},
  pages={165-174},
}

@inproceedings{chibane2020neural,
    title = {Neural Unsigned Distance Fields for Implicit Function Learning},
    author = {Chibane, Julian and Mir, Aymen and Pons-Moll, Gerard},
    booktitle = {Advances in Neural Information Processing Systems ({NeurIPS})},
    year = {2020},
}

@article{wang2023neural,
author = {Wang, Zixiong and Zhang, Yunxiao and Xu, Rui and Zhang, Fan and Wang, Peng-Shuai and Chen, Shuangmin and Xin, Shiqing and Wang, Wenping and Tu, Changhe},
title = {Neural-Singular-Hessian: Implicit Neural Representation of Unoriented Point Clouds by Enforcing Singular Hessian},
year = {2023},
volume = {42},
number = {6},
journal = {ACM Trans. Graph.},
}

@InProceedings{Lorensen1987,
  author    = {Lorensen, William E. and Cline, Harvey E.},
  booktitle = {Proceedings of the 14th Annual Conference on Computer Graphics and Interactive Techniques},
  title     = {Marching Cubes: A High Resolution 3D Surface Construction Algorithm},
  year      = {1987},
  pages     = {163--169},
  series    = {SIGGRAPH '87},
  numpages  = {7},
}

@article{VIPSS,
  author       = {Zhiyang Huang and
                  Nathan Carr and
                  Tao Ju},
  title        = {Variational implicit point set surfaces},
  journal      = {{ACM} Trans. Graph.},
  volume       = {38},
  number       = {4},
  pages        = {124:1--124:13},
  year         = {2019},
}

@INPROCEEDINGS{venkatesh2021csp,
   author={Venkatesh, Rahul and Karmali, Tejan and Sharma, Sarthak and Ghosh, Aurobrata and Babu, R. Venkatesh and Jeni, L\'{a}szl\'{o} A. and Singh, Maneesh},
  booktitle={2021 IEEE/CVF International Conference on Computer Vision (ICCV)}, 
  title={Deep Implicit Surface Point Prediction Networks}, 
  year={2021},
  volume={},
  number={},
  pages={12633-12642},
  doi={10.1109/ICCV48922.2021.01242}
}

@inproceedings{NeuralPull,
    title = {{Neural-Pull:} Learning Signed Distance Functions from Point Clouds by Learning to Pull Space onto Surfaces},
    author = {Ma, Baorui and Han, Zhizhong and Liu, Yu-Shen and Zwicker, Matthias},
    booktitle = {Proceedings of the International Conference on Machine Learning (ICML)
    },
    year = {2021}
}

@inproceedings{DBLP:conf/iccv/RenHCHW23,
  author       = {Siyu Ren and
                  Junhui Hou and
                  Xiaodong Chen and
                  Ying He and
                  Wenping Wang},
  title        = {GeoUDF: Surface Reconstruction from 3D Point Clouds via Geometry-guided
                  Distance Representation},
  booktitle    = {{ICCV}},
  pages        = {14168--14178},
  year         = {2023},
}

@article{hou2023dcudf,
author = {Hou, Fei and Chen, Xuhui and Wang, Wencheng and Qin, Hong and He, Ying},
title = {Robust Zero Level-Set Extraction from Unsigned Distance Fields Based on Double Covering},
year = {2023},
volume = {42},
number = {6},
journal = {ACM Trans. Graph.},
articleno = {245},
numpages = {15},
}

@InProceedings{Fainstein2024DUDF,
    author    = {Fainstein, Miguel and Siless, Viviana and Iarussi, Emmanuel},
    title     = {DUDF: Differentiable Unsigned Distance Fields with Hyperbolic Scaling},
    booktitle = {Proceedings of the IEEE/CVF Conference on Computer Vision and Pattern Recognition (CVPR)},
    month     = {June},
    year      = {2024},
    pages     = {4484-4493}
}

@ARTICLE{yang2025monge,
    author={Yang, Chuanxiang and Zhou, Yuanfeng and Wei, Guangshun and Ma, Long and Hou, Junhui and Liu, Yuan and Wang, Wenping},
    journal={IEEE Transactions on Pattern Analysis and Machine Intelligence},
    title={Monge-Ampere Regularization for Learning Arbitrary Shapes from Point Clouds}, 
    year={2025},
    volume={47},
    number={8},
    pages={1-15},
}

@inproceedings{Wang22HSDF,
    author = {Wang, Li and Yang, Jie and Chen, Wei-Kai and Meng, Xiao-Xu and Yang, Bo and Li, Jin-Tao and Gao, Lin},
    title = {HSDF: Hybrid Sign and Distance Field for Modeling Surfaces with Arbitrary Topologies },
    booktitle={Neural Information Processing Systems (NeurIPS)},
    year = {2022},
}

@inproceedings{ye2022gifs,
  title={Gifs: Neural implicit function for general shape representation},
  author={Ye, Jianglong and Chen, Yuntao and Wang, Naiyan and Wang, Xiaolong},
  booktitle={2022 IEEE/CVF Conference on Computer Vision and Pattern Recognition (CVPR)},
  pages={12819--12829},
  year={2022},
  organization={IEEE}
}

@inproceedings{
kong2026metricphase,
title={Metric{\textemdash}Phase Fields: Decoupling Distance and Sign for Thin-Structure Reconstruction from Unoriented Point Clouds},
author={Jiayi Kong and Xuhui Chen and Chen Zong and Fei Hou and Junhui Hou and Wenping Wang and Ying He},
booktitle={ICML 2026},
year={2026},
}

@inproceedings{alexa2001point,
  title={Point set surfaces},
  author={Alexa, Marc and Behr, Johannes and Cohen-Or, Daniel and Fleishman, Shachar and Levin, David and Silva, Claudio T},
  booktitle={Proceedings Visualization, 2001. VIS'01.},
  pages={21--29},
  year={2001},
  organization={IEEE}
}

@article{kolluri2008provably,
  title={Provably good moving least squares},
  author={Kolluri, Ravikrishna},
  journal={ACM Transactions on Algorithms (TALG)},
  volume={4},
  number={2},
  pages={1--25},
  year={2008},
  publisher={ACM New York, NY, USA}
}

@article{xia2025variational,
  title={Variational surface reconstruction using natural neighbors},
  author={Xia, Jianjun and Ju, Tao},
  journal={ACM Transactions on Graphics (TOG)},
  volume={44},
  number={4},
  pages={1--19},
  year={2025},
  publisher={ACM New York, NY, USA}
}

@article{mello2025neural,
  title={Neural vector fields for implicit surface representation and inference},
  author={Mello Rella, Edoardo and Chhatkuli, Ajad and Konukoglu, Ender and Van Gool, Luc},
  journal={International Journal of Computer Vision},
  volume={133},
  number={4},
  pages={1855--1878},
  year={2025},
  publisher={Springer}
}

@inproceedings{yang2023neural,
  title={Neural vector fields: Implicit representation by explicit learning},
  author={Yang, Xianghui and Lin, Guosheng and Chen, Zhenghao and Zhou, Luping},
  booktitle={2023 IEEE/CVF Conference on Computer Vision and Pattern Recognition (CVPR)},
  pages={16727--16738},
  year={2023},
  organization={IEEE}
}

@article{10.1109/TVCG.2023.3318085,
author = {Tian, Hui and Zhu, Chenyang and Shi, Yifei and Xu, Kai},
title = {SuperUDF: Self-Supervised UDF Estimation for Surface Reconstruction},
year = {2024},
volume = {30},
number = {9},
journal = {IEEE Transactions on Visualization and Computer Graphics},
pages = {5965--5975},
numpages = {11}
}

@article{liu2025unsupervised,
  title={Unsupervised point cloud reconstruction via recurrent multi-step moving strategy},
  author={Liu, Zheng and Zhang, Jianjun and Zhang, Ming and Ke, Runze and Yu, Chengcheng and Liu, Ligang},
  journal={IEEE Transactions on Multimedia},
  volume={28},
  pages={972--984},
  year={2025},
  publisher={IEEE}
}

\appendix

\newpage

Appendices~\ref{app:soft-set-proof}--\ref{app:objective-convexity-proof} prove the propositions and main theorem. Appendix~\ref{sec:hm} describes confidence-guided heat diffusion, scalar integration, and surface extraction. Appendix~\ref{app:benchmark} describes the benchmark used for evaluation and comparison. Finally, Appendix~\ref{app:additional-results} presents ablation studies, compares PNF with VAD, presents additional visual and quantitative results, and discusses the limitations. 

\section{Proof of Proposition~\ref{prop:soft-set-convex}}
\label{app:soft-set-proof}

We establish convexity, compactness, the convex-hull identity,
and the characterization of extreme points.

\paragraph{Convexity.}
Let \(\mathbf M,\mathbf N\in\mathcal P_3^{\mathrm{cvx}}\)
and \(t\in[0,1]\). The matrix
\(\mathbf L=(1-t)\mathbf M+t\mathbf N\) is symmetric.
For every \(\mathbf x\in\mathbb R^3\),
\[
\mathbf x^\top\mathbf L\mathbf x
=
(1-t)\mathbf x^\top\mathbf M\mathbf x
+
t\mathbf x^\top\mathbf N\mathbf x
\geq0,
\]
so \(\mathbf L\succeq0\). Moreover,
\[
\operatorname{tr}(\mathbf L)
=
(1-t)\operatorname{tr}(\mathbf M)
+
t\operatorname{tr}(\mathbf N)
=
1.
\]
Thus, \(\mathbf L\in\mathcal P_3^{\mathrm{cvx}}\), proving convexity.

\paragraph{Compactness.}
The positive-semidefinite cone and the unit-trace constraint are closed,
so \(\mathcal P_3^{\mathrm{cvx}}\) is closed in
\(\operatorname{Sym}(3)\).
For any \(\mathbf M\in\mathcal P_3^{\mathrm{cvx}}\),
its eigenvalues are nonnegative and sum to one. Hence
\[
\|\mathbf M\|_F^2
=
\sum_{r=1}^{3}\lambda_r^2
\leq
\left(\sum_{r=1}^{3}\lambda_r\right)^2
=
1.
\]
The set is therefore bounded. Since
\(\operatorname{Sym}(3)\) is finite-dimensional,
closedness and boundedness imply compactness.

\paragraph{Convex-hull identity.}
By the spectral theorem, every
\(\mathbf M\in\mathcal P_3^{\mathrm{cvx}}\) can be written as
\[
\mathbf M
=
\sum_{r=1}^{3}\lambda_r\mathbf e_r\mathbf e_r^\top,
\qquad
\lambda_r\geq0,
\qquad
\sum_{r=1}^{3}\lambda_r=1,
\]
where the eigenvectors form an orthonormal basis.
Each \(\mathbf e_r\mathbf e_r^\top\) belongs to \(\mathcal P_3\).
Thus,
\(\mathcal P_3^{\mathrm{cvx}}
\subseteq\operatorname{conv}(\mathcal P_3)\).

Conversely, \(\mathcal P_3\subseteq\mathcal P_3^{\mathrm{cvx}}\),
and the latter set is convex. It therefore contains every convex
combination of hard projectors. Hence
\[
\mathcal P_3^{\mathrm{cvx}}
=
\operatorname{conv}(\mathcal P_3).
\]

\paragraph{Extreme points.}
First, let
\(\mathbf P=\mathbf n\mathbf n^\top\in\mathcal P_3\)
and suppose
\[
\mathbf P=(1-t)\mathbf M+t\mathbf N,
\qquad
\mathbf M,\mathbf N\in\mathcal P_3^{\mathrm{cvx}},
\qquad
0<t<1.
\]
For every \(\mathbf x\perp\mathbf n\),
\[
0
=
\mathbf x^\top\mathbf P\mathbf x
=
(1-t)\mathbf x^\top\mathbf M\mathbf x
+
t\mathbf x^\top\mathbf N\mathbf x.
\]
Both terms on the right are nonnegative, so each must vanish.

For a positive-semidefinite matrix \(\mathbf A\),
\(\mathbf x^\top\mathbf A\mathbf x=0\) implies
\(\mathbf A\mathbf x=0\), because
\[
\mathbf x^\top\mathbf A\mathbf x
=
\|\mathbf A^{1/2}\mathbf x\|_2^2.
\]
Consequently, \(\mathbf M\) and \(\mathbf N\) vanish on
\(\mathbf n^\perp\). By symmetry, their ranges are contained in
\(\operatorname{span}\{\mathbf n\}\).
Their unit traces then imply
\(\mathbf M=\mathbf N=\mathbf P\).
Thus, every hard projector is an extreme point.

Conversely, suppose
\(\mathbf M\in\mathcal P_3^{\mathrm{cvx}}\) has rank greater than one.
At least two eigenvalues are positive; denote them by
\(\lambda_1,\lambda_2>0\), with corresponding orthonormal eigenvectors
\(\mathbf e_1,\mathbf e_2\).
Choose \(0<\varepsilon<\min\{\lambda_1,\lambda_2\}\) and define
\[
\mathbf M_\pm
:=
\mathbf M
\pm\varepsilon
\left(
\mathbf e_1\mathbf e_1^\top
-
\mathbf e_2\mathbf e_2^\top
\right).
\]
Both matrices are positive semidefinite, symmetric, and have unit trace.
They are distinct and satisfy
\(\mathbf M=(\mathbf M_++\mathbf M_-)/2\).
Therefore, \(\mathbf M\) is not extreme.
The extreme points of \(\mathcal P_3^{\mathrm{cvx}}\)
are exactly the hard projectors.
\hfill\(\square\)

\section{Proof of Proposition~\ref{prop:nearest-hard-projector}}
\label{app:nearest-hard-proof}

Every hard projector has the form
$\mathbf P=\mathbf n\mathbf n^\top$, with $\|\mathbf n\|_2=1$. Using
\(\mathbf P^2=\mathbf P\) and
\(\operatorname{tr}(\mathbf P)=1\), we obtain
\begin{align*}
\|\mathbf M-\mathbf P\|_F^2
&=
\operatorname{tr}
\left(
(\mathbf M-\mathbf P)^2
\right)\\
&=
\operatorname{tr}(\mathbf M^2)
+
\operatorname{tr}(\mathbf P^2)
-
2\operatorname{tr}(\mathbf M\mathbf P)\\
&=
\operatorname{tr}(\mathbf M^2)
+
1
-
2\mathbf n^\top\mathbf M\mathbf n.
\end{align*}
The first two terms are independent of \(\mathbf n\). Finding
the nearest hard projector is therefore equivalent to maximizing
\(\mathbf n^\top\mathbf M\mathbf n\) over unit vectors.

By the Rayleigh--Ritz theorem,
\[
\max_{\|\mathbf n\|_2=1}
\mathbf n^\top\mathbf M\mathbf n
=
\lambda_1,
\]
where \(\lambda_1\) is the largest eigenvalue of \(\mathbf M\). The
maximizers are exactly the unit vectors in the leading eigenspace.
Hence the minimum squared distance is $1+\operatorname{tr}(\mathbf M^2)-2\lambda_1$, attained by 
$\operatorname{Hard}(\mathbf M)$. 
The minimizing projector is unique exactly when the leading eigenspace
is one-dimensional, equivalently when
\(\lambda_1>\lambda_2\).

Finally, the eigenvalues of $\mathbf{M}$ are nonnegative and sum to one, so
\[
\operatorname{tr}(\mathbf M^2)
=
\sum_{r=1}^{3}\lambda_r^2
\leq
\left(
\sum_{r=1}^{3}\lambda_r
\right)^2
=
1.
\]
Equality holds exactly when one eigenvalue is one and the others are zero, which characterizes the hard projectors.

The Cauchy--Schwarz inequality gives
\[
1
=
\left(
\sum_{r=1}^{3}\lambda_r
\right)^2
\leq
3\sum_{r=1}^{3}\lambda_r^2.
\]
Therefore, $\operatorname{tr}(\mathbf M^2)\geq\frac13$, with equality exactly when all three eigenvalues equal \(1/3\), that is, when
\(\mathbf M=\mathbf I/3\).
\hfill\(\square\)

\section{Proof of Theorem~\ref{thm:pnf-convexity}}
\label{app:objective-convexity-proof}

By Proposition~\ref{prop:soft-set-convex},
\(\mathcal P_3^{\mathrm{cvx}}\) is compact and convex.
It is nonempty because it contains \(\mathbf I/3\).
Its finite Cartesian product
\((\mathcal P_3^{\mathrm{cvx}})^N\)
is therefore nonempty, compact, and convex.

All local scatter matrices, soft-PCA priors, graph edges, unit edge
directions, and weights are fixed.
The objective is a finite sum of linear and quadratic functions of the
tensor entries, so it is continuous.

To prove strong convexity, we use
\begin{equation}
\|(1-t)\mathbf a+t\mathbf b\|^2
=
(1-t)\|\mathbf a\|^2
+
t\|\mathbf b\|^2
-
t(1-t)\|\mathbf a-\mathbf b\|^2,
\qquad 0\leq t\leq1.
\label{eq:squared-norm-identity}
\end{equation}
This identity holds for the Euclidean and Frobenius norms and for
the ordinary square of a scalar.

Let \(\mathbf A\) and \(\mathbf B\) be feasible tensor fields, and set
\(\mathbf L_t=(1-t)\mathbf A+t\mathbf B\).
The field \(\mathbf L_t\) is feasible by convexity.
Since the fitting energy is linear,
\begin{equation}
E_{\mathrm{tan}}(\mathbf L_t)
=
(1-t)E_{\mathrm{tan}}(\mathbf A)
+
tE_{\mathrm{tan}}(\mathbf B).
\label{eq:tangency-linear-combination}
\end{equation}

Applying Equation~(\ref{eq:squared-norm-identity}) to each anchoring
term yields
\begin{equation}
E_{\mathrm{anc}}(\mathbf L_t)
=(1-t)E_{\mathrm{anc}}(\mathbf A)
+tE_{\mathrm{anc}}(\mathbf B)
-\frac{t(1-t)}{2}\sum_{i=1}^{N}
\rho_i\|\mathbf A_i-\mathbf B_i\|_F^2.
\label{eq:anchor-strong-convexity}
\end{equation}
Because
\(\rho_i\geq\rho_{\min}:=\min_{1\leq i\leq N}\rho_i>0\),
\begin{equation}
E_{\mathrm{anc}}(\mathbf L_t)
\leq (1-t)E_{\mathrm{anc}}(\mathbf A)
+
tE_{\mathrm{anc}}(\mathbf B)
-
\frac{\rho_{\min}}{2}
t(1-t)\|\mathbf A-\mathbf B\|_{\mathbb F}^2.
\label{eq:anchor-lower-curvature}
\end{equation}

For every edge \(\{i,j\}\in E\), the expressions
\[
\mathbf M_i-\mathbf M_j,\qquad
(\mathbf M_i+\mathbf M_j)\widehat{\mathbf d}_{ij},\qquad
\widehat{\mathbf d}_{ij}^{\top}
(\mathbf M_i-\mathbf M_j)\widehat{\mathbf d}_{ij}
\]
are linear in the tensor field.
Equation~(\ref{eq:squared-norm-identity}) therefore implies that
their squared norms or scalar squares are convex.
Since all overlap weights are nonnegative,
\begin{equation}
E_{\mathrm{ov}}(\mathbf L_t)
\leq
(1-t)E_{\mathrm{ov}}(\mathbf A)
+
tE_{\mathrm{ov}}(\mathbf B).
\label{eq:overlap-convexity}
\end{equation}

Combining the fitting, anchoring, and overlap inequalities gives
\begin{equation}
E_{\mathrm{PNF}}(\mathbf L_t)
\leq{}
(1-t)E_{\mathrm{PNF}}(\mathbf A)
+
tE_{\mathrm{PNF}}(\mathbf B)
-
\frac{\rho_{\min}}{2}
t(1-t)\|\mathbf A-\mathbf B\|_{\mathbb F}^2.
\label{eq:pnf-strong-convexity}
\end{equation}
This is precisely \(\rho_{\min}\)-strong convexity with respect
to the block Frobenius norm.

Continuity on the nonempty compact feasible set guarantees the existence
of a global minimizer.
To prove uniqueness, suppose that two distinct feasible fields
\(\mathbf A\) and \(\mathbf B\) both attain the minimum \(E^\star\).
Their midpoint is feasible, and Equation~(\ref{eq:pnf-strong-convexity})
with \(t=1/2\) gives
\[
E_{\mathrm{PNF}}\left(\frac{\mathbf A+\mathbf B}{2}\right)
\leq
E^\star
-
\frac{\rho_{\min}}{8}
\|\mathbf A-\mathbf B\|_{\mathbb F}^2
<
E^\star.
\]
This contradicts minimality. The global minimizer is therefore unique.
\hfill\(\square\)

\section{Confidence-Guided Heat Diffusion for UDF Construction}
\label{sec:hm}

Stage~II converts the normal axes estimated by PNF into a global scalar distance approximation. Following the geometry-first strategy of VAD~\citep{DBLP:journals/corr/abs-2510-12524}, we propagate directional information through the ambient domain before scalar integration. PNF additionally supplies the eigengap confidence, which controls source selection and weighting. Thus, Stage~I optimizes soft tensors, whereas Stage~II uses their decoded axes and confidence values.

\paragraph{Bidirectional sources.}
For each sample $\mathbf p_i$, choose either unit representative $\mathbf n_i$ of the decoded axis and place two sources at $\mathbf p_i\pm\varepsilon\mathbf n_i$, where $\varepsilon>0$ is a small offset. The sources carry directions $+\mathbf n_i$ and $-\mathbf n_i$, pointing away from the local plane on their respective sides. Changing the representative to $-\mathbf{n}_i$ exchanges the two sources but leaves the construction unchanged. Placing the opposite directions at distinct positions avoids the cancellation that would occur if they were averaged at the same point. The construction therefore provides two-sided directional information without globally orienting the input normals.

\paragraph{Confidence weighting and truncation.}
We propagate the paired directional sources through the ambient domain using heat-kernel
evaluations, with contributions weighted by the eigengap
confidence \(c_i\). Some outliers retain small but nonzero confidence
values despite unreliable normal-axis estimates, allowing their
source contributions to influence the propagated field. To limit this residual influence, we define the \textit{effective} source weight as follows:
\[
\widetilde c_i
=
\begin{cases}
0, & c_i<0.1,\\
c_i, & c_i\geq0.1.
\end{cases}
\]
The threshold $0.1$ was
selected empirically from the tested examples. Both sources in each
pair receive the same weight, preserving invariance to reversal of the chosen normal representation. 

The truncation suppresses weak directional evidence rather than classifying or removing input points.
It can suppress ambiguous axes near valid non-manifold junctions as well as those associated with corrupted observations. By itself, it neither removes the original samples nor modifies their scalar-value constraints. Table~\ref{tab:confidence-ablation} compares uniform source weighting, raw confidence weighting, and the truncation scheme.

The propagated field can be evaluated at arbitrary spatial positions,
rather than only at reconstruction-grid vertices. However, cancellation
and ambiguity can persist near the surface, junctions, and medial
structures; the field therefore need not coincide with an exact UDF
gradient, which may be undefined at such locations. Likewise,
\(c_i\) measures how distinctly the optimized tensor favors a normal
axis, not whether that axis agrees with the unknown ground-truth normal.

\paragraph{Scalar integration.}
Following heat-based distance reconstruction~\citep{feng2024heat,DBLP:journals/corr/abs-2510-12524}, we use the propagated directions to guide scalar integration on  a discretized ambient domain. Continuous directional queries provide local guidance, while the discrete solve assembles a global distance approximation.
Source weighting and truncation are distinct from scalar-value
constraints: suppressing a directional source does not automatically
relax a zero-value constraint imposed at the corresponding sample.

\paragraph{Surface extraction.} We use the double-covering approach of DCUDF~\citep{hou2023dcudf}. Using a small positive isovalue, Marching Cubes~\citep{Lorensen1987} first extracts an isosurface around the target. DCUDF then jointly optimizes its vertices toward the zero level set, using field values at vertices and triangle centroids together with geometric regularization. In our implementation, vertex updates use
continuous queries of the propagated directional field rather than
relying solely on numerical derivatives of the discretized UDF.
For orientable manifold targets, DCUDF can separate the double cover
into a single-layer mesh; otherwise, it retains a double-layered
geometric approximation.

\begin{table*}[!hp] \centering 
\caption{Effect of neighborhood graph construction on bidirectional normal estimation. We evaluate PNF using symmetrized $k$-NN, radius-based, and Voronoi-adjacency graphs while keeping the input, objective weights, and optimization settings unchanged. Each graph remains fixed during optimization. We report mean absolute cosine similarity (higher is better) and mean angular error in degrees (lower is better) between estimated and reference normal axes. For non-manifold models, evaluation is restricted to junction neighborhoods rather than the entire surface. PNF maintains comparable accuracy across the three graph constructions and outperforms local PCA in both metrics on the evaluated thin structures and non-manifold models.  } 
\label{tab:graph-ablation} \setlength{\tabcolsep}{5pt} \renewcommand{\arraystretch}{1.2} \begin{tabular}{llcc} \toprule \textbf{Model features} & \textbf{Graph type} & \textbf{Normal accuracy ($\uparrow$)} & \textbf{Mean angular error ($^\circ, \downarrow$)} \\ 

 \midrule \multirow{4}{*}{Thin structures} & $k$-NN & 0.9661 & 3.673 \\ & Radius & 0.9657 & 3.877 \\ & Voronoi & 0.9651 & 4.236 \\ & PCA & 0.9479 & 9.132 \\ \midrule \multirow{4}{*}{Non-manifold models} & $k$-NN & 0.9540 & 10.040 \\ & Radius & 0.9444 & 10.976 \\ & Voronoi & 0.9483 & 10.690 \\ & PCA & 0.9184 & 14.822\\ \bottomrule \end{tabular} \end{table*}

\section{Benchmark}
\label{app:benchmark}

Our benchmark contains 60 point-cloud models: 10 medial-geometry models, 10 garments, 5 models with complex topology, 15 everyday objects with open or non-manifold surfaces, and 20 indoor scenes. The benchmark point clouds contain approximately 50,000--150,000 samples. Together, they cover thin sheets, open boundaries, closely spaced surface layers, non-manifold junctions, and geometrically complex scenes. 

These categories test complementary aspects of reconstruction. Thin-structure models require nearby surface sheets to remain distinct, while open and non-manifold models test reconstruction around boundaries and junctions.  Models with complex topology and scene-level geometry provide additional configurations in which local surface evidence must be combined into a coherent distance field. 

For each model, we evaluate five input conditions: clean samples,
positional noise at levels of \(0.3\%\) and \(0.8\%\), and outlier contamination at levels of \(2\%\) and \(5\%\). The same set of 60 models is used across all conditions, and all methods are evaluated under the same corruption settings. Noise perturbs
the local surface geometry, whereas outliers introduce samples that do not belong to the underlying surface. These experiments assess reconstruction accuracy and robustness across the
different geometric categories and input conditions.

\section{Additional Results and Discussion}
\label{app:additional-results}

This section examines graph construction, normal estimation, and confidence guidance through controlled comparisons. It also compares PNF with VAD, reports the computational cost of the pipeline, and discusses its limitations. 

\subsection{Ablation of Normal Estimation and Confidence Guidance}
\label{app:reconstruction-ablations}

The main experiments evaluate the complete reconstruction pipeline.
Table~\ref{tab:confidence-ablation} isolates the effects of the normal source and the use of
PNF confidence while keeping the reconstruction back-end fixed.
Table~\ref{tab:graph-ablation} specifies a separate comparison of neighborhood graph
constructions.

\paragraph{Graph construction.}
We compare symmetrized \(k\)-nearest-neighbor, radius-based, and
Voronoi-adjacency graphs while keeping the input, objective weights,
and reconstruction settings fixed. Each graph is constructed before
optimization and remains fixed throughout the solve. We also compare
PCA and PNF using identical local neighborhoods, helping distinguish
the contribution of joint tensor optimization from that of neighborhood
selection.

Table~\ref{tab:graph-ablation} reports the results on the four thin-structure models and three non-manifold models in Figures~\ref{fig:knn-noise-vis} and~\ref{fig:noise+outlier}, respectively. PNF maintains comparable normal-estimation
accuracy across the three graph constructions on these densely and
approximately uniformly sampled inputs. This suggests that its joint
optimization is not strongly dependent on a particular neighborhood
rule under the tested sampling conditions.

\paragraph{Confidence guidance.}
Table~\ref{tab:confidence-ablation} compares PCA, VAD, and PNF normal sources under uniform weighting, then evaluates two additional PNF variants: direct weighting by the raw eigengap confidence and confidence truncation as described in Appendix~\ref{sec:hm}. The input contains both noise and outliers, and the
reconstruction back-end and remaining settings are held fixed.
These comparisons separate the choice of normal source from the
subsequent use of confidence.

With uniform weights, PNF improves CD and F-score relative to both
PCA and VAD, although VAD has a slightly lower HD95.
Applying raw confidence weights to PNF improves CD from \(6.98\)
to \(6.44\) and F-score from \(95.65\%\) to \(96.06\%\),
while HD95 increases slightly from \(26.07\) to \(26.53\).
Confidence truncation achieves the best reported result in all three
metrics: CD \(3.73\), HD95 \(6.22\), and F-score \(99.20\%\).
These results support the use of source truncation in the evaluated
corruption setting, rather than assuming that raw confidence weighting
alone improves every aspect of reconstruction.


\begin{table*}[!hp]\centering
\caption{Ablation of normal sources and confidence guidance. Inputs contain both noise and outliers, while the reconstruction back-end and remaining settings are fixed. PCA, VAD, and the first PNF variant use uniform source weights $c_i\equiv 1$. The remaining PNF variants use raw eigengap weights $c_i$ or thresholded weights $\widetilde{c}_i$ as defined in Appendix~\ref{sec:hm}. CD and HD95 are reported in units of $10^{-3}$; F-score is reported as a percentage at distance tolerance $0.01$. The best result in each metric is shown in \textbf{bold}.}
\label{tab:confidence-ablation}
\setlength{\tabcolsep}{6pt}\renewcommand{\arraystretch}{1.2}
\begin{tabular}{ll ccc}
\toprule
\textbf{Normal} & \textbf{Confidence} & \textbf{CD ($\downarrow$)}  & \textbf{HD95 ($\downarrow$)}  & \textbf{F-score ($\uparrow$)}\\
\midrule
PCA & Uniform $c_i\equiv 1$              & 8.98 & 34.03 & 89.36\\
VAD & Uniform $c_i\equiv 1$              & 7.04 & 25.49 & 95.11\\
\midrule
PNF & Uniform $c_i\equiv 1$              & 6.98 & 26.07 & 95.65\\
PNF & Raw eigengap $c_i$      & 6.44 & 26.53 & 96.06\\
PNF & Thresholded eigengap $\widetilde{c}_i$          & \textbf{3.73} & \textbf{6.22} & \textbf{99.20}\\
\bottomrule
\end{tabular}
\end{table*}

\begin{figure*}[t]
\centering
\includegraphics[width=\linewidth]{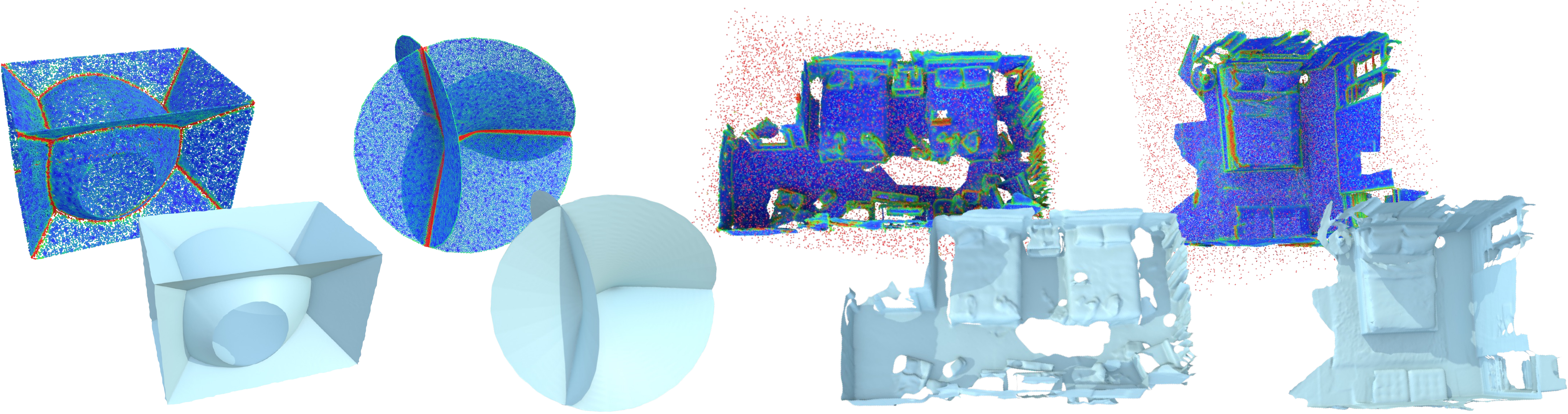}
\vspace{-0.1in}
\caption{Normal-axis confidence as a geometric cue. The upper row shows input points colored by the eigengap confidence of the optimized PNF tensors; the lower row shows the corresponding reconstructions. The left examples contain non-manifold junctions, and the right examples contain outliers. Low confidence indicates weak axial preference and can highlight candidate junction regions or unreliable directional estimates.}
\label{fig:confidence-applications}
\end{figure*}

\subsection{Robustness to Input Density}
\begin{figure*}[t]
\centering
\includegraphics[width=\linewidth]{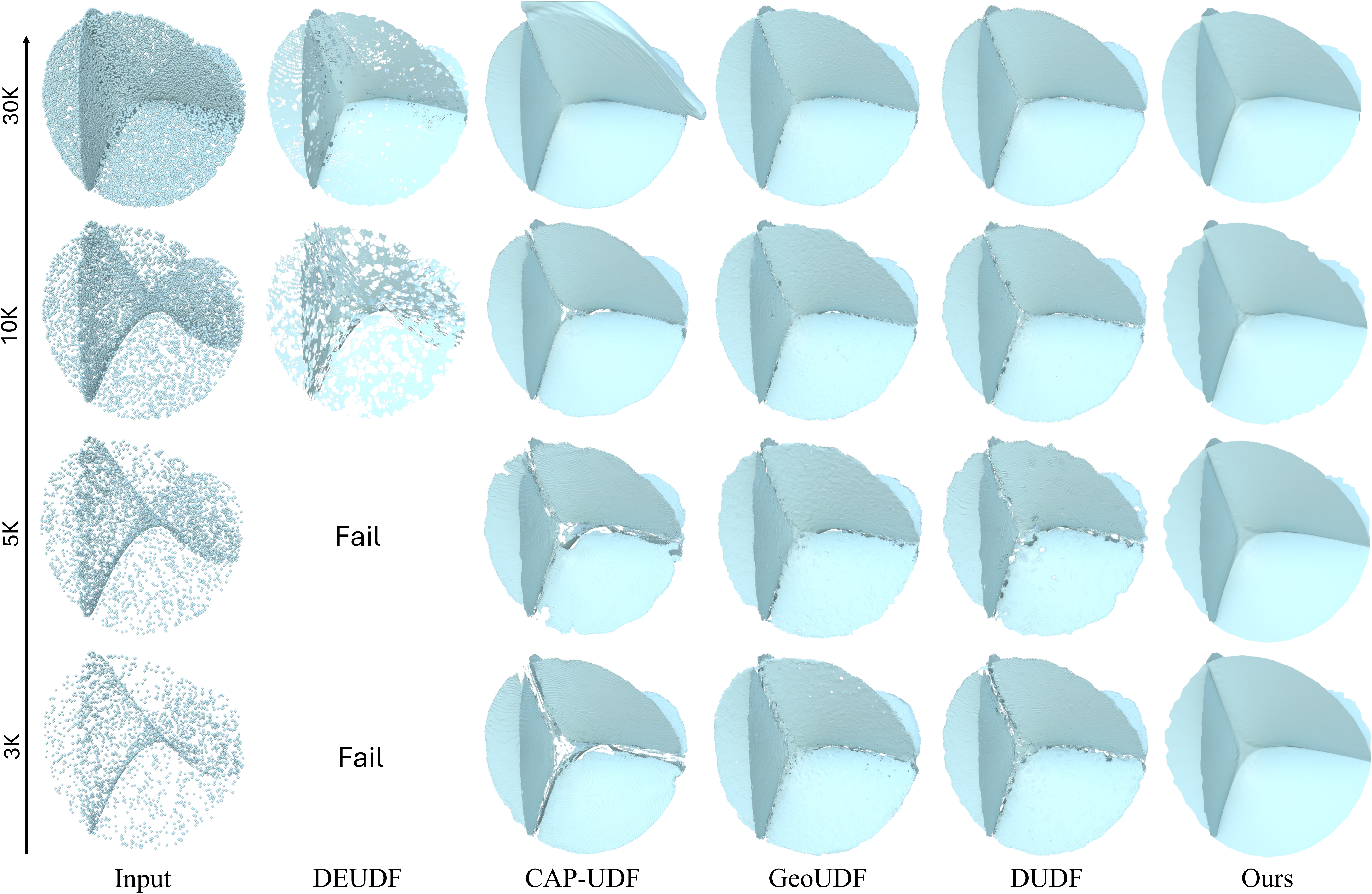}
\caption{Qualitative comparison under varying point-cloud densities. With the input size decreasing from 30K to 3K points, competing methods gradually suffer from geometric distortions or reconstruction failures, whereas our method consistently produces stable and coherent surfaces, demonstrating robustness to sparse point-cloud observations.}
\label{fig:varydensity}
\end{figure*}

We further evaluate the robustness of different reconstruction methods to variations in input point density (Figure~\ref{fig:varydensity}). The network-based methods included in our experiments generally exhibit degraded reconstruction quality as the number of input points decreases, and may even fail to recover a valid surface mesh under sufficiently sparse inputs. In contrast, our optimization-based approach consistently produces stable reconstructions across a wide range of input point counts. Even with substantially fewer input points, PNF continues to recover coherent surface geometry while preserving fine structures and complex topological configurations. These results indicate that our method is less dependent on a specific sampling density and remains effective under varying degrees of input sparsity.

\subsection{Comparison with VAD}
\label{subsec:relation-to-vad}
PNF and VAD~\citep{DBLP:journals/corr/abs-2510-12524} share a geometry-first strategy: estimate bidirectional normals from point positions, propagate the resulting directional information, and integrate it into a global UDF. VAD also represents optimized bidirectional normals by rank-one tensors \(\mathbf n_i\mathbf n_i^\top\) during tensor diffusion. 
 
 Despite their shared sign-invariant tensor representation, the two methods differ fundamentally in how the bidirectional normal field is formulated and optimized. 
 
 VAD directly optimizes bidirectional normal vectors by enforcing consistency between local projection-distance fields across Voronoi bisectors. Its Voronoi diagram determines both which fields interact and where their value and gradient discrepancies are evaluated. The Voronoi structure is therefore an integral part of its normal-optimization framework, rather than merely a neighborhood data structure. The resulting optimization is solved iteratively without a reported global-optimality guarantee. 
 
 PNF instead uses normal-axis projectors as optimization variables and relaxes the rank-one constraint to obtain a convex set of positive-semidefinite, unit-trace tensors. It couples these tensors over a fixed neighborhood graph by comparing the Hessians, midpoint gradients, and midpoint values of local quadratic models. A Voronoi construction is not required, and positive anchoring weights guarantee a unique global minimizer. The optimized tensor spectrum also provides axial confidence, which
PNF uses for directional-source selection and weighting during
reconstruction. Thus, PNF combines convex soft-tensor optimization
with confidence-guided propagation while retaining the high-level
geometry-first pipeline. 

The reported results also favor the complete PNF pipeline over VAD:
PNF achieves lower directed CD and HD under every corruption condition
in Table~\ref{tab:overall-benchmark}, and lower global CD, junction-region CD, and HD95 on all
three non-manifold models in Table~\ref{tab:nonmanifold-eval}.

\subsection{Computational Cost and Runtime}
\label{subsec:runtime}

The pipeline consists of PNF-based normal-axis estimation followed by UDF construction and surface extraction. During PNF estimation, the neighborhood graph is constructed once and remains fixed. With fixed-size tensor updates at each point and a bounded amount of work per graph edge, each optimization iteration requires $O(N+|E|)$ operations, where $N$ and $|E|$ denote the number of input points and graph edges, respectively. We adopt a simple gradient descent solver and use 500 optimization iterations per point cloud. 

On the evaluated non-manifold point clouds, PNF optimization takes \(8.299\,\mathrm{s}\) on average, excluding graph construction. Among the tested graph constructions, Voronoi adjacency has the largest average construction time, \(0.0272\,\mathrm{s}\), at the evaluated point-cloud scale.

Stage~II uses confidence-guided heat diffusion and scalar integration to construct the UDF, followed by DCUDF surface extraction. Field computation and surface extraction take \(  329.269\,\mathrm{s}\) and \(35.036\,\mathrm{s}\) on average, respectively. We report these components separately to distinguish the cost of PNF estimation from that of downstream reconstruction. On these examples, field computation accounts for most of the reported time.

\subsection{Limitations}
\label{app:limitations}

PNF's global-optimality guarantee applies to the fixed-graph soft-tensor optimization in Stage~I, not to exact surface recovery. Reconstruction
remains dependent on graph connectivity, local geometric evidence, and numerical discretization. Surface extraction may retain a double-layered mesh whose geometry approximates the target without reproducing its topology. Recovering an appropriate single-layer representation requires additional post-processing, particularly for non-manifold or non-orientable targets.



\begin{table*}[t]
\centering
\caption{\textbf{Reconstruction accuracy on non-manifold models.}
Global and junction-region metrics are reported for three synthetic
shapes. Distances are evaluated from the reference geometry to the
reconstruction. CD and HD95 are reported in units of \(10^{-3}\),
and recall as a percentage. The junction-band width and recall
tolerance are \(2\%\) and \(0.5\%\) of the reference bounding-box
diagonal, respectively. The best result for each metric and shape
is shown in \textbf{bold}, including ties. GeoUDF uses the PUGeo-Net
\(\times16\) upsampling variant~\citep{PUGeo}.
DEUDF was not included in the table, since it produced severely degraded reconstructions on these examples under the evaluated settings. The main reason is its reliance on locally estimated PCA normals for gradient alignment. Near non-manifold junctions, neighborhoods containing multiple surface sheets can yield unreliable normal axes and misleading directional constraints. Although PNF also uses local PCA information, it treats the resulting tensors as soft priors rather than final normal-axis estimates. Crucially, PNF jointly optimizes the entire normal-axis field over a connectivity graph, allowing individual estimates to be refined through compatibility with neighboring geometric evidence. This global coupling enables information from well-supported neighborhoods to help resolve ambiguous local estimates, while the soft representation retains competing directional preferences where ambiguity persists. Thus, PNF's advantage lies not in avoiding local PCA, but in reconciling its evidence through a globally coupled, strongly convex normal-estimation problem.}
\label{tab:nonmanifold-eval}

\small
\setlength{\tabcolsep}{4pt}
\renewcommand{\arraystretch}{1.15}

\begin{tabular}{@{}clrrrr@{}}
\toprule
\textbf{Model}
& \textbf{Method}
& \textbf{Global CD} \(\downarrow\)
& \textbf{Junction CD} \(\downarrow\)
& \textbf{Junction Recall} \(\uparrow\)
& \textbf{HD95} \(\downarrow\) \\
\midrule

\multirow{6}{*}{%
  \rotatebox[origin=c]{90}{%
    \shortstack[c]{%
      \footnotesize\emph{Cross-junction}\\
      \footnotesize (3,000 points)%
    }%
  }%
}
& CAP-UDF
& 0.148 & 0.356 & 99.77 & 0.299 \\
& GeoUDF
& 0.050 & 0.224 & \textbf{100.00} & 0.117 \\
& DUDF
& 0.200 & 1.221 & 99.75 & 0.597 \\
& VAD
& 0.038 & 0.321 & \textbf{100.00} & 0.204 \\
& PCA+HM
& 0.066 & 0.598 & \textbf{100.00} & 0.380 \\
& \textbf{Ours}
& \textbf{0.014} & \textbf{0.115}
& \textbf{100.00} & \textbf{0.071} \\
\midrule

\multirow{6}{*}{%
  \rotatebox[origin=c]{90}{%
    \shortstack[c]{%
      \footnotesize\emph{Multi-junction}\\
      \footnotesize (4,431 points)%
    }%
  }%
}
& CAP-UDF
& 0.808 & 3.189 & 85.17 & 5.546 \\
& GeoUDF
& 0.082 & 0.289 & \textbf{100.00} & 0.210 \\
& DUDF
& 0.591 & 2.066 & 96.49 & 2.580 \\
& VAD
& 0.090 & 0.371 & \textbf{100.00} & 0.417 \\
& PCA+HM
& 0.188 & 0.789 & \textbf{100.00} & 1.021 \\
& \textbf{Ours}
& \textbf{0.031} & \textbf{0.123}
& \textbf{100.00} & \textbf{0.156} \\
\midrule

\multirow{6}{*}{%
  \rotatebox[origin=c]{90}{%
    \shortstack[c]{%
      \footnotesize\emph{Henneberg surface}\\
      \footnotesize (2,911 points)%
    }%
  }%
}
& CAP-UDF
& 1.072 & 3.675 & 85.59 & 5.899 \\
& GeoUDF
& 0.535 & 1.359 & 99.85 & 1.830 \\
& DUDF
& 0.562 & 2.020 & 97.07 & 2.638 \\
& VAD
& 0.475 & 1.795 & 99.30 & 2.483 \\
& PCA+HM
& 0.456 & 1.758 & 99.17 & 2.567 \\
& \textbf{Ours}
& \textbf{0.384} & \textbf{1.255}
& \textbf{100.00} & \textbf{1.688} \\

\bottomrule
\end{tabular}
\end{table*}

\begin{figure*}[htb]
    \centering
    \includegraphics[width=0.98\linewidth]{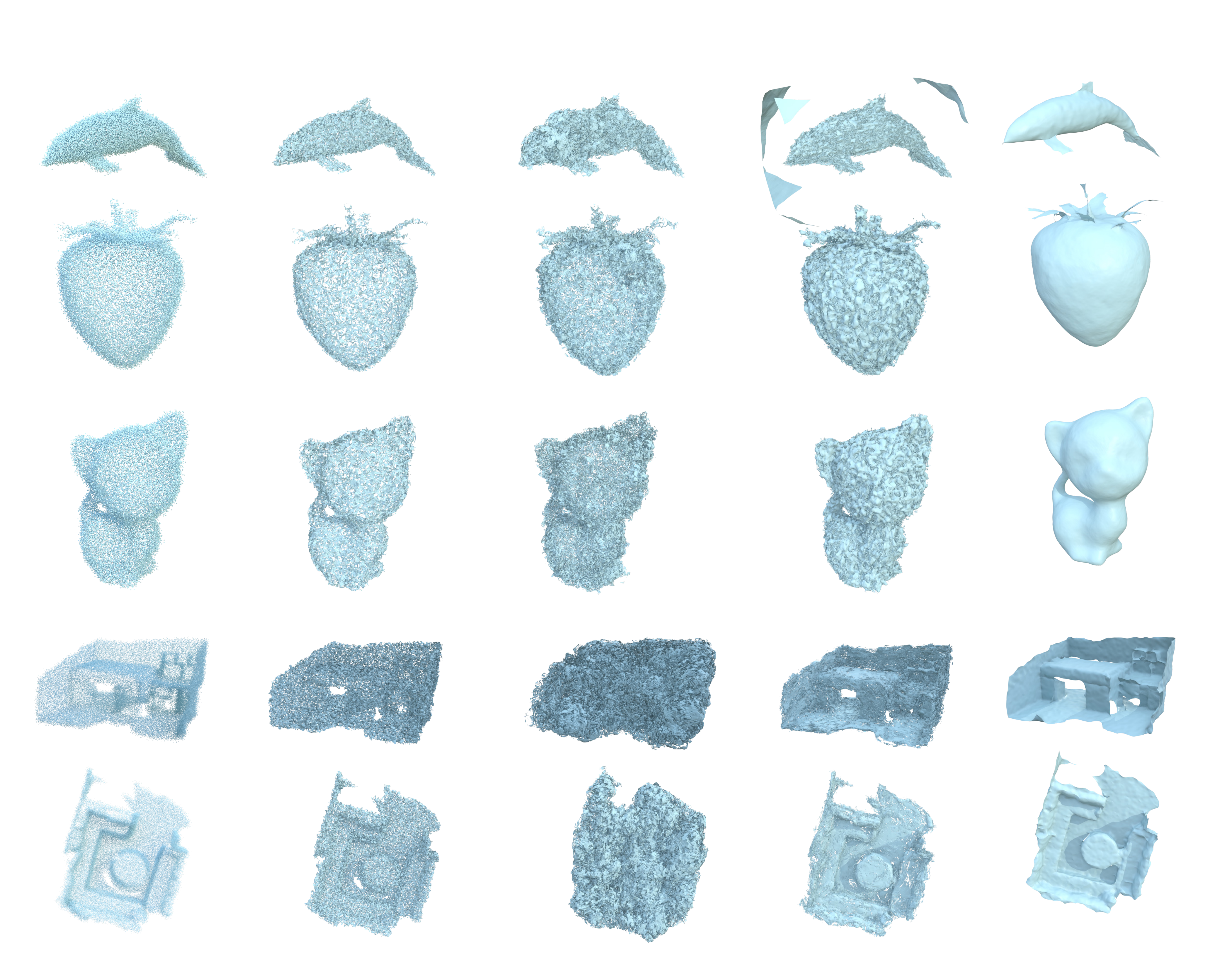}
    \makebox[0.17\linewidth][c]{Input}
     \makebox[0.17\linewidth][c]{GeoUDF}
       \makebox[0.17\linewidth][c]{DUDF}
     \makebox[0.17\linewidth][c]{CAP-UDF} 
     \makebox[0.17\linewidth][c]{Ours}
    \caption{Reconstruction from noisy inputs. The input point clouds are corrupted by  $0.8\%$ Gaussian noise. Several competing methods produce fragmented patches or irregular surfaces, whereas PNF yields smoother, more coherent reconstructions in these examples. Table~\ref{tab:overall-benchmark} reports quantitative results on the 60-model benchmark.}
    
    \label{fig:noise}
\end{figure*}

\begin{figure*}[htb]
    \centering
    \begin{minipage}{0.8\textwidth}
    \centering
    \includegraphics[width=0.98\linewidth]{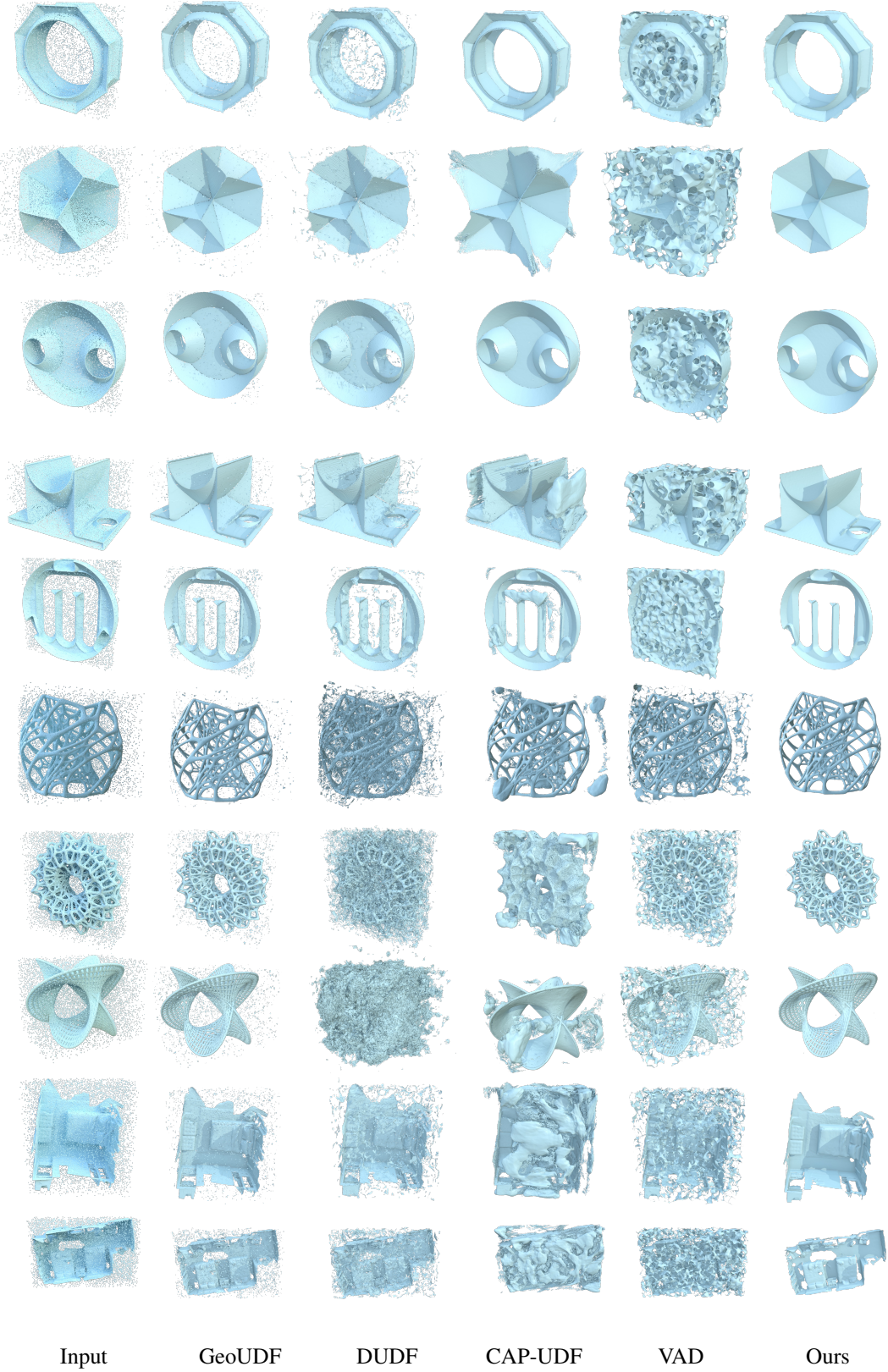}
    \makebox[0.17\linewidth][c]{Input}
     \makebox[0.15\linewidth][c]{GeoUDF}
       \makebox[0.15\linewidth][c]{DUDF}
     \makebox[0.15\linewidth][c]{CAP-UDF} 
          \makebox[0.15\linewidth][c]{VAD} 
     \makebox[0.15\linewidth][c]{Ours}
    \end{minipage}
    \caption{Reconstruction from outlier-contaminated inputs. The point clouds contain $5\%$ outliers sampled within the bounding box. Several competing methods produce spurious patches or surface distortions, whereas PNF yields more coherent reconstructions with fewer visible artifacts in these examples. Table~\ref{tab:overall-benchmark} reports quantitative results on the 60-model benchmark. }
    
    \label{fig:outlier}
\end{figure*}

\end{document}